\documentclass[10pt,twocolumn,letterpaper]{article}
\usepackage[OT1]{fontenc}
\usepackage[pagenumbers]{cvpr}

\usepackage{cuted} %< for strip environment

\usepackage{currfile} %< for \currfilebase macro

\usepackage{caption} %< for "\captionof" commands (teaser)
\usepackage{amsmath}
\usepackage{booktabs,multirow,graphicx,amssymb}
\usepackage{xcolor}
\usepackage{placeins}
\usepackage{needspace}
\usepackage{stfloats}
\usepackage{longtable,overpic}
\definecolor{cvprblue}{rgb}{0.21,0.49,0.74}
\usepackage[pagebackref,breaklinks,colorlinks,allcolors=cvprblue]{hyperref}

\title{UAV-CinePilot: Can Embodied Foundation Models Master Autonomous Aerial Cinematography?}

\author{%
Xiangdong Li\textsuperscript{1},
Jianli Sun\textsuperscript{2},
Sibing Wei\textsuperscript{3},
Siyao Liu\textsuperscript{4}\\
Changtao Miao\textsuperscript{5},
Jian Di\textsuperscript{5},
Deyi Ji\textsuperscript{5},
Yonglin Tian\textsuperscript{2,}\thanks{Corresponding author: \texttt{yonglin.tian@ia.ac.cn}.}\\[4pt]
{\small \textsuperscript{1}Institute of Computing Technology, Chinese Academy of Sciences}\\
{\small \textsuperscript{2}Institute of Automation, Chinese Academy of Sciences}\\
{\small \textsuperscript{3}Beijing Institute of Technology\quad
\textsuperscript{4}Hebei University}\\
{\small \textsuperscript{5}University of Science and Technology of China}
}

\begin{document}
\maketitle

\begin{abstract}
Language-guided aerial cinematography requires an unmanned aerial vehicle (UAV) to translate natural-language instructions into coordinated flight and camera motions while maintaining desired visual relationships with a subject. Unlike point-goal navigation or generic target tracking, autonomous filming demands simultaneous control over target-relative geometry, camera trajectory, and image framing. To systematically benchmark end-to-end embodied policies on this task, we introduce \textbf{UAV-CinePilot}, a closed-loop aerial cinematography benchmark. We define a structured photography task space with five factors: \emph{target state}, \emph{relative distance}, \emph{relative azimuth}, \emph{relative elevation}, and \emph{framing}. Within this space, we instantiate four foundational shot families into 41 concrete cinematography semantics. Using Unreal Engine 5.5 and AirSim, we collect 9,840 language-annotated filming episodes across ten simulation environments and 149 subject instances. We further propose six complementary metrics to decouple motion correctness, task progress, subject retention, image framing, and extraneous motion. Closed-loop evaluations of nine representative policies across 2,460 test cases reveal a clear capability gap: while current embodied foundation models and action policies execute basic camera motions and keep subjects broadly in view, they struggle to sustain requested framing during motion, adapt to moving targets, and coordinate multiple simultaneous filming constraints.
\end{abstract}

\begin{figure*}[t]
    \centering
    \includegraphics[width=\textwidth]{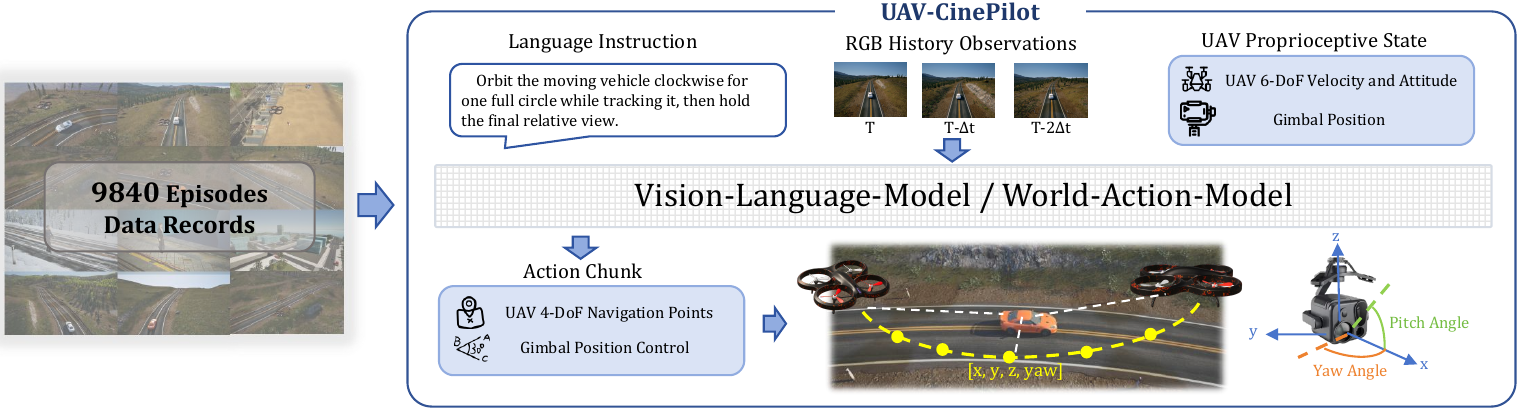}
    \caption{\textbf{Overview of the aerial cinematography benchmark.}
    Given a natural-language filming instruction, temporally ordered onboard RGB observations,
    and UAV proprioceptive states, an embodied policy predicts action chunks to jointly control
    UAV flight and gimbal orientation. The benchmark evaluates whether closed-loop execution
    realizes the requested camera--subject relationship using a dataset of 9,840 filming episodes.}
    \label{fig:benchmark_overview}
\end{figure*}

% Required packages:
% \usepackage{booktabs,multirow,amssymb}

\newcommand{\yes}{\ensuremath{\checkmark}}
\newcommand{\no}{\ensuremath{\times}}
\newcommand{\partialcap}{\ensuremath{\triangle}}

\begin{table*}[t]
\centering
\caption{\textbf{Comparison with representative UAV tasks.}
T-Lang and M-Lang indicate language-based target and
motion specification, respectively.
E2E denotes a learned policy mapping visual observations
and language instructions directly to actions, whereas
% State the visual-feedback criterion directly.
{V-Loop denotes action updates based on visual feedback
of the subject during execution.}
Dist., Azim., and Elev. indicate explicitly specified
target-relative distance, azimuth, and elevation constraints;
Framing denotes image-space subject positioning rather
than overall cinematic quality.
\yes: explicitly addressed;
\partialcap: partially addressed or not a primary
evaluation objective;
\no: not explicitly addressed.
S/M: static/moving subjects.
Symbols characterize the primary task scope rather
than all possible capabilities of each method.}
\label{tab:related_task_comparison}

\scriptsize
\setlength{\tabcolsep}{4pt}
\renewcommand{\arraystretch}{1.12}

\begin{tabular}{lccccccccc}
\toprule
& \multicolumn{2}{c}{Language}
& Target
& \multirow{2}{*}{E2E}
& \multirow{2}{*}{V-Loop}
& \multicolumn{4}{c}{Cinematic Constraints} \\
\cmidrule(lr){2-3}
\cmidrule(lr){7-10}
Method
& T-Lang & M-Lang & State
& & & Dist. & Azim. & Elev. & Framing \\
\midrule

\multicolumn{10}{l}{
\textit{Autonomous Aerial Cinematography}} \\

Galvane et al.~\cite{
galvane2017directingcinematographicdrones}
& \no & \partialcap & S/M
& \no & \partialcap
& \yes & \yes & \yes & \yes \\

ACDC~\cite{
lin2025agenticaerialcinematographydialogue}
& \yes & \yes & S
& \no & \partialcap
& \partialcap & \partialcap & \partialcap & \yes \\

DiffusionCinema~\cite{
serpiva2026diffusioncinematexttoaerialcinematography}
& \partialcap & \yes & S
& \no & \no
& \partialcap & \yes & \partialcap & \partialcap \\

\midrule
\multicolumn{10}{l}{
\textit{UAV Vision-Language Navigation / Control}} \\

AerialVLN~\cite{Liu_2023_ICCV}
& \yes & \yes & S
% Learned Seq2Seq/CMA baselines map visual and language inputs directly to actions.
& {\yes} & \yes
& \partialcap & \partialcap & \partialcap & \no \\

OpenUAV~\cite{
wang2024realisticuavvisionlanguagenavigation}
& \yes & \yes & S
& \partialcap & \yes
& \partialcap & \partialcap & \partialcap & \no \\

AutoFly~\cite{
sun2026autoflyvisionlanguageactionmodeluav}
& \partialcap & \yes & S
& \yes & \yes
& \partialcap & \partialcap & \partialcap & \no \\

UAV-Flow~\cite{
wang2025uavflowcolosseorealworldbenchmark}
& \yes & \yes & S
& \yes & \yes
& \yes & \partialcap & \partialcap & \no \\

\midrule
\multicolumn{10}{l}{
\textit{Embodied UAV Tracking}} \\

UAV-Track VLA~\cite{
zhang2026uavtrackvlaembodiedaerial}
& \yes & \partialcap & M
& \yes & \yes
% The task explicitly specifies distance and checks distance-aware tracking success.
& {\yes} & \partialcap & \partialcap & \partialcap \\

CosFly-VLA~\cite{
ren2026cosflyvlaspatiallyawarevisionlanguageaction}
& \yes & \partialcap & M
& \yes & \yes
& \yes & \partialcap & \partialcap & \partialcap \\

\midrule
\textbf{UAV-CinePilot (Ours)}
& \yes & \yes & S/M
& \yes & \yes
& \yes & \yes & \yes & \yes \\

\bottomrule
\end{tabular}
\end{table*}

\section{Introduction}
\label{sec:introduction}

Language-guided aerial cinematography requires unmanned aerial vehicles (UAVs) to execute filming tasks specified by natural-language instructions. Unlike point-goal navigation, which primarily aims to reach a destination, autonomous filming requires a UAV to maintain desired visual relationships with a subject throughout a shot. For example, an instruction such as \emph{``Follow the moving car from its rear-left side while gradually ascending and pulling away''} specifies target-relative geometry, camera trajectory, and the intended evolution of the captured view. Realizing such instructions demands joint visual grounding, spatial reasoning, and continuous flight and gimbal control.

Embodied foundation models, including vision-language-action (VLA) and world-action architectures, offer a unified framework for this task~\cite{pmlr-v229-zitkovich23a,BlackK-RSS-25}. Rather than relying on separately engineered perception, planning, and control modules, end-to-end policies directly map onboard visual history and language instructions to physical actions. In aerial filming, language grounding supports diverse shot descriptions, while closed-loop visual feedback enables continuous adaptation as subjects and viewpoints change. Figure~\ref{fig:benchmark_overview} illustrates this language-conditioned UAV and gimbal control interface.

Existing UAV benchmarks have advanced language-guided
navigation~\cite{Liu_2023_ICCV,
wang2024realisticuavvisionlanguagenavigation,
sun2026autoflyvisionlanguageactionmodeluav},
fine-grained flight control~\cite{
wang2025uavflowcolosseorealworldbenchmark},
embodied aerial tracking~\cite{
zhang2026uavtrackvlaembodiedaerial,
ren2026cosflyvlaspatiallyawarevisionlanguageaction},
and aerial manipulation~\cite{pmlr-v306-sun26an}.
As summarized in Table~\ref{tab:related_task_comparison},
these tasks primarily emphasize goal-directed navigation,
motion execution, or target following, without systematically
evaluating the joint satisfaction of cinematic motion and
image-space framing constraints.
Existing aerial cinematography systems~\cite{
galvane2017directingcinematographicdrones,
bonatti2019autonomousaerialcinematographyunstructured,
lin2025agenticaerialcinematographydialogue,
serpiva2026diffusioncinematexttoaerialcinematography}
explicitly address viewpoint planning and framing, but
typically rely on modular pipelines that separate cinematic
planning from low-level control.
In contrast, language-guided aerial cinematography requires
embodied policies to jointly interpret target and motion
instructions, adapt to static or moving subjects through
closed-loop visual feedback, and coordinate UAV and gimbal
actions to satisfy target-relative distance, azimuth,
elevation, and image-space framing constraints.
A dedicated benchmark for systematically evaluating
these coupled capabilities in end-to-end embodied policies
remains lacking.

% Clarify how the research questions relate to prior tracking and cinematography work.
{In this work, we investigate three core questions: \emph{(i)} Can embodied policies ground natural-language filming instructions into the required camera--subject geometry and visual composition? \emph{(ii)} Can they execute coordinated UAV and gimbal actions while maintaining consistent visual feedback during dynamic shots? \emph{(iii)} Can they coordinate multiple simultaneous constraints in compositional filming instructions? Prior works explore related capabilities through language-guided tracking and trajectory planning~\cite{zhang2026uavtrackvlaembodiedaerial,lin2025agenticaerialcinematographydialogue,serpiva2026diffusioncinematexttoaerialcinematography}. In contrast, we focus on jointly evaluating end-to-end cinematography policies across language grounding, coordinated control, and multi-constraint satisfaction. Answering these questions requires a benchmark centered on cinematic execution rather than destination-reaching alone.}

To this end, we introduce \textbf{UAV-CinePilot}, a benchmark for language-conditioned, closed-loop aerial cinematography. We define a structured photography task space comprising five factors: \emph{target state}, \emph{relative distance}, \emph{relative azimuth}, \emph{relative elevation}, and \emph{framing}. Combining four foundational shot families---\emph{Follow}, \emph{Orbit}, \emph{Elevation}, and \emph{Fly-over}---with compatible spatial and framing constraints yields 41 concrete cinematography semantics, covering both elementary and compositional shots. Across ten simulation maps and 149 filming-subject instances, we collect 9,840 language-annotated filming episodes with diverse target motions and spatial configurations. {We generate these episodes using task-conditioned expert controllers and verify them through automated consistency checks and manual video inspection.} We evaluate closed-loop execution using six complementary metrics spanning \emph{motion correctness}, \emph{task progress}, \emph{subject retention}, \emph{image framing}, and \emph{extraneous motion}. Evaluating nine representative policies across 2,460 test cases shows that current models execute basic motions and keep subjects broadly in frame, yet exhibit sharp drops in specified framing, moving-target tracking, and multi-constraint coordination.

Our main contributions are summarized as follows:
\begin{itemize}
    \item We introduce \textbf{UAV-CinePilot}, a closed-loop benchmark and 9,840-episode dataset for evaluating end-to-end embodied policies on language-guided UAV cinematography.
    \item We establish a structured task taxonomy of 41 elementary and compositional filming semantics, paired with six fine-grained metrics that decouple 3D motion execution from image-space composition.
    \item We systematically benchmark nine representative policies across VLA, world-action, and classic action architectures, pinpointing key bottlenecks in dynamic tracking, precise framing, and multi-constraint coordination.
\end{itemize}

\section{Related Work}
\label{sec:related_work}

\subsection{UAV Vision-Language Navigation and Control}
Language-conditioned UAV research has evolved from aerial navigation benchmarks, such as AerialVLN~\cite{Liu_2023_ICCV} and OpenUAV~\cite{wang2024realisticuavvisionlanguagenavigation}, toward end-to-end flight policies. AutoFly~\cite{sun2026autoflyvisionlanguageactionmodeluav} explores VLA learning for autonomous navigation, while UAV-Flow~\cite{wang2025uavflowcolosseorealworldbenchmark} benchmarks fine-grained, language-guided flight control. Embodied tracking frameworks, including TrackVLA~\cite{pmlr-v305-wang25f}, UAV-Track VLA~\cite{zhang2026uavtrackvlaembodiedaerial}, and CosFly-VLA~\cite{ren2026cosflyvlaspatiallyawarevisionlanguageaction}, demonstrate semantic target grounding and continuous visual tracking. AIR-VLA~\cite{pmlr-v306-sun26an} further extends aerial VLA evaluation to manipulation. Rather than focusing on navigation, tracking, or manipulation alone, our benchmark evaluates the coupled execution of camera trajectory and visual composition specified by cinematic instructions.

\subsection{Autonomous Aerial Cinematography}
Classical aerial cinematography systems rely on explicit geometric constraints, viewpoint optimization, and trajectory planning to generate camera motions. Galvane et al.~\cite{galvane2017directingcinematographicdrones} introduce cinematographic viewpoint representations with physically feasible planning, while Bonatti et al.~\cite{bonatti2019autonomousaerialcinematographyunstructured} integrate visual tracking, motion optimization, and learned shot selection. Recent language-driven frameworks, such as ACDC~\cite{lin2025agenticaerialcinematographydialogue} and DiffusionCinema~\cite{serpiva2026diffusioncinematexttoaerialcinematography}, translate textual directing cues into executable filming trajectories. Unlike task-specific modular pipelines, end-to-end embodied models directly unify visual perception and action generation. Our benchmark provides a standardized testbed to measure these end-to-end capabilities in closed-loop filming.

\subsection{Embodied AI and Cinematography Benchmarks}
EmbodiedBench~\cite{pmlr-v267-yang25f} and MetaVQA~\cite{Wang_2025_CVPR} assess embodied reasoning, scene understanding, and interactive decision-making, whereas MM-UAVBench~\cite{dai2025mmuavbenchmultimodallargelanguage} targets multimodal perception in low-altitude UAV scenarios. CineTechBench~\cite{wang2025cinetechbenchbenchmarkcinematographictechnique} evaluates knowledge and synthesis of cinematographic techniques in video generation. In contrast to static visual question answering, open-loop video generation, or generic flight benchmarks, our work directly measures whether an embodied policy can realize requested cinematic shots through closed-loop UAV and gimbal control.

\section{Benchmark Design and Construction}
\label{sec:benchmark}

\subsection{Task Formulation}
\label{sec:task_formulation}

We formulate aerial cinematography as a language-conditioned, closed-loop control task. An embodied policy must execute specified camera motions while maintaining the desired visual relationship with a filming subject. Given a natural-language instruction $\ell$, onboard visual observations, and UAV proprioceptive states, the policy predicts actions that jointly control the UAV and its gimbal. Unlike point-goal navigation, autonomous filming requires temporally consistent cinematographic behavior rather than reaching a single destination. Multiple valid trajectories can therefore satisfy the same instruction as long as they fulfill the specified spatial and visual constraints.

Formally, an embodied policy $\pi_\theta$ predicts a sequence of future actions:
\begin{equation}
    \mathbf{A}_t =
    \pi_\theta(\ell,\mathcal{I}_t,\mathbf{s}_t),
    \qquad
    \mathbf{A}_t =
    \{\mathbf{a}_{t+k}\}_{k=0}^{H-1},
    \label{eq:cinematography_policy}
\end{equation}
where $\mathcal{I}_t$ denotes the visual observation history, $\mathbf{s}_t$ represents UAV proprioceptive states, and $H$ is the prediction horizon. We define a six-dimensional incremental action space:
\begin{equation}
    \mathbf{a}_t =
    [\Delta x_b,\Delta y_b,\Delta z_b,
    \Delta\psi_b,\Delta\theta_g,\Delta\psi_g],
    \label{eq:uav_action}
\end{equation}
where $[\Delta x_b,\Delta y_b,\Delta z_b]$ denotes translation in a heading-aligned body frame, and $[\Delta\psi_b,\Delta\theta_g,\Delta\psi_g]$ specifies UAV yaw, gimbal pitch, and relative gimbal yaw increments.

The standard observation interface provides three temporally ordered RGB frames and UAV proprioception, including motion states, camera orientation, and previous control commands. Ground-truth target positions, velocities, and relative geometry are used solely for data collection and evaluation, and remain hidden from the policy during inference.

\subsection{Cinematography Task Taxonomy}
\label{sec:task_taxonomy}

\begin{figure*}[t]
    \centering
    \includegraphics[width=\textwidth]{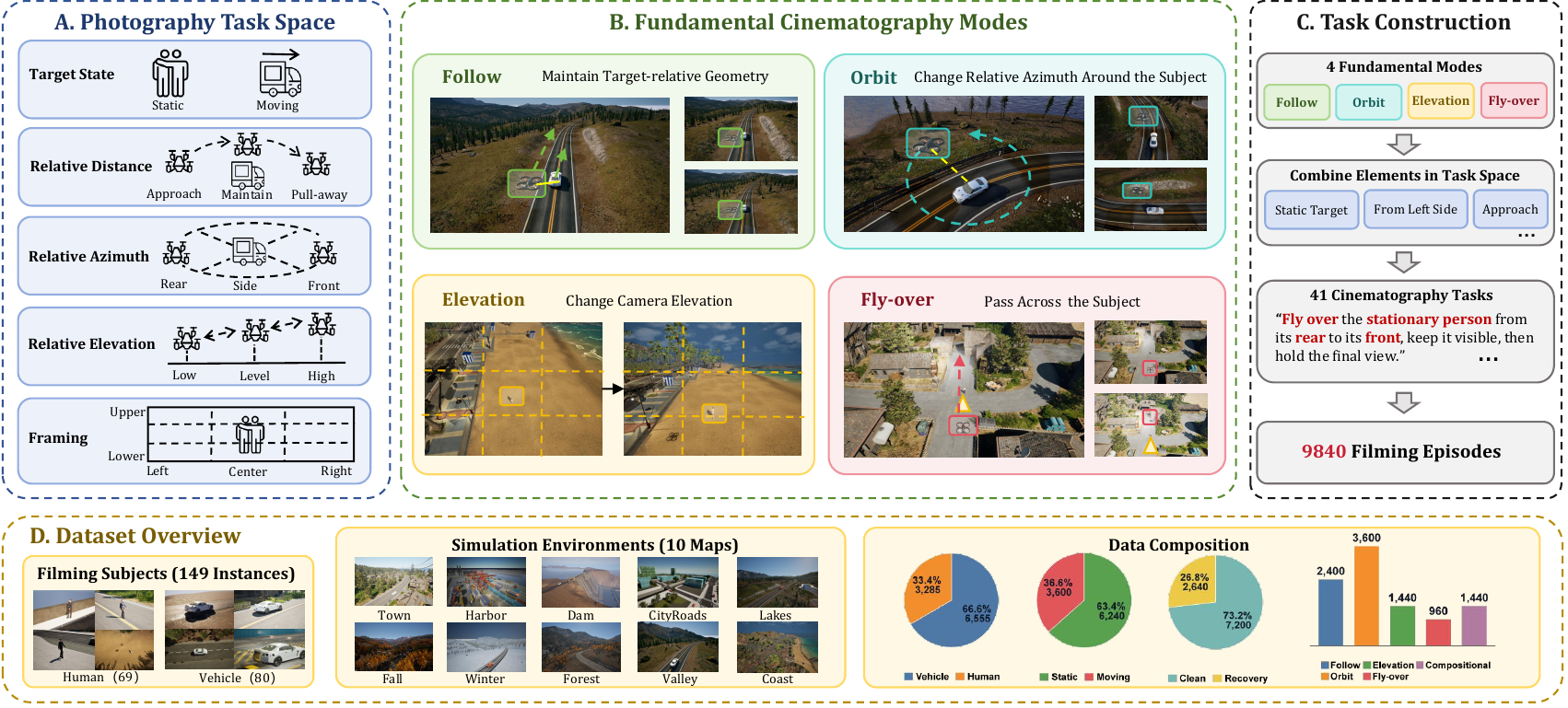}
    \caption{\textbf{Cinematography task taxonomy and dataset overview.}
    \textbf{(A)} Five task-space factors specify target state, relative distance,
    relative azimuth, relative elevation, and framing. \textbf{(B)} Representative shots from
    the four foundational cinematography families. \textbf{(C)} Task construction
    combines compatible filming requirements into 41 concrete cinematography
    semantics across 9,840 episodes. \textbf{(D)} Dataset distribution across ten simulation
    environments and 149 subject instances, broken down by subject category,
    target motion, demonstration type, and task family.}
    \label{fig:dataset_overview}
\end{figure*}

\noindent\textbf{Photography Task Space.}
We define a structured photography task space to capture the core spatial and visual requirements of aerial filming across five factors: \emph{target state}, \emph{relative distance}, \emph{relative azimuth}, \emph{relative elevation}, and \emph{framing}. Target state distinguishes static and moving subjects. Relative distance specifies whether the camera maintains, approaches, or pulls away from the subject. Relative azimuth and elevation define the target-relative viewing geometry, while framing specifies the desired subject position in the image plane. These factors define cinematographic objectives independently of specific subject identities (Fig.~\ref{fig:dataset_overview}A).

\noindent\textbf{Foundational Cinematography Families.}
Based on this task space, we organize aerial filming into four foundational shot families: \emph{Follow}, \emph{Orbit}, \emph{Elevation}, and \emph{Fly-over}. \emph{Follow} maintains a prescribed relative configuration with a moving subject. \emph{Orbit} produces continuous relative azimuth rotation around the subject. \emph{Elevation} emphasizes vertical camera movement, optionally paired with distance changes. \emph{Fly-over} passes directly across the subject to transition between opposing viewpoints. Each family incorporates compatible task-space factors to cover diverse filming behaviors while avoiding physically inconsistent combinations.

\noindent\textbf{Concrete Cinematography Semantics.}
Combining the four foundational families with compatible task-space factors yields 41 concrete cinematography semantics. Of these, 35 are elementary semantics---ten \emph{Follow}, 15 \emph{Orbit}, six \emph{Elevation}, and four \emph{Fly-over}---and six are compositional semantics. An elementary semantic specifies a single primary behavior, such as following a moving subject from directly behind. Compositional semantics combine multiple filming requirements simultaneously, such as maintaining a side-following viewpoint while gradually pulling away. Each semantic is instantiated across diverse subjects, environments, and motion parameters to construct individual filming episodes (Fig.~\ref{fig:dataset_overview}B--C).

We further categorize all 41 semantics into two execution regimes. \emph{Continuous filming} tasks (eight semantics) require maintaining a specified camera--subject relationship throughout the filming interval. \emph{Finite-motion} tasks (33 semantics) require completing a prescribed camera trajectory and holding its terminal configuration. This partition encompasses both sustained visual tracking and goal-directed camera movements.
% Point to the complete semantic catalogue without expanding the main taxonomy.
{The complete semantic catalogue and analysis groups are listed in Appendix~\ref{app:semantic_groups}.}

\subsection{Simulation Environments and Target Diversity}
\label{sec:environments}

We construct the benchmark in Unreal Engine 5.5 and AirSim{~\cite{shah2017airsimhighfidelityvisualphysical}} to simulate UAV flight dynamics and onboard RGB sensing. The camera operates with fixed intrinsics, achieving all cinematographic motions through UAV translation, body yaw, and gimbal control.

To ensure broad visual and geometric diversity, we build filming tasks across ten simulation maps spanning coastal landscapes, urban roads, harbors, dams, lakeside forests, mountains, canyons, and seasonal environments. Each map is configured with compatible subjects, motion paths, and collision-free flight regions. Every map covers all 41 concrete cinematography semantics, ensuring consistent task distributions across distinct visual scenes.

The dataset includes 149 map-specific subject instances, comprising 80 vehicles and 69 human subjects (Fig.~\ref{fig:dataset_overview}D). Subjects are assigned according to task requirements and trajectory feasibility: \emph{Follow} tasks use moving subjects, while the remaining tasks use compatible static or moving subjects as specified by their cinematography semantics. This design introduces rich variation in target appearance, scale, motion, and background geometry while preserving consistent task semantics.

\subsection{Data Collection and Trajectory Diversification}
\label{sec:data_collection}

\begin{figure*}[t]
    \centering
    \includegraphics[width=\textwidth]{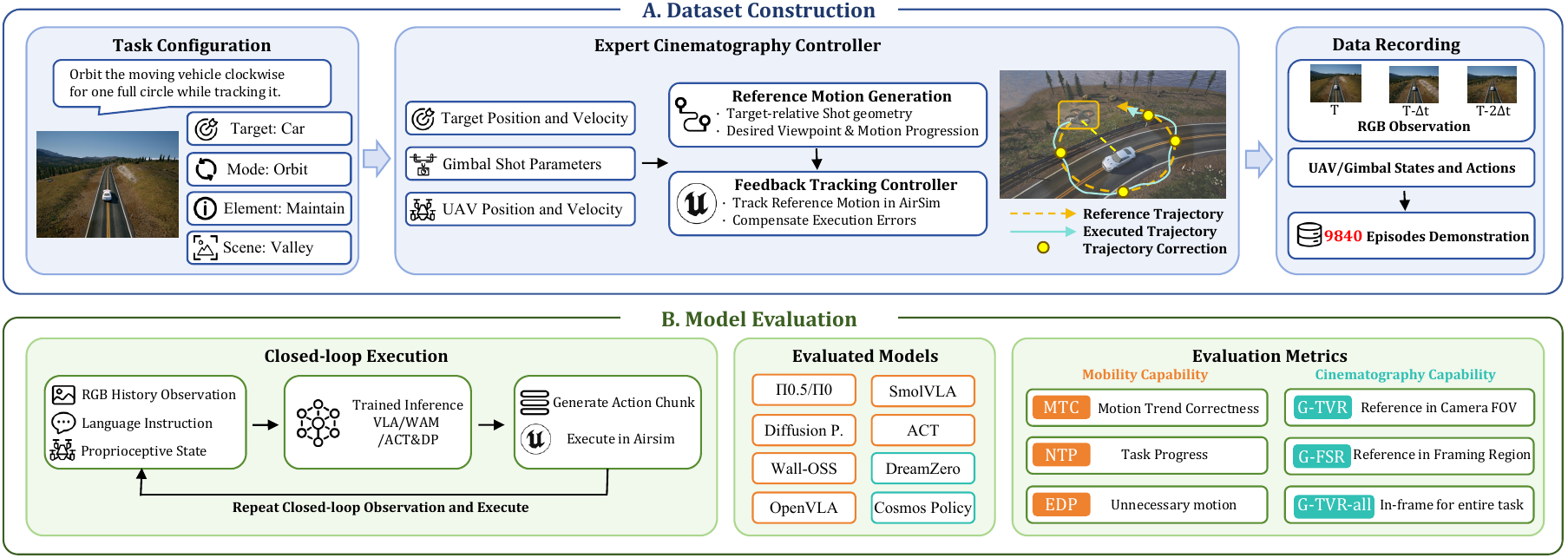}
    \caption{\textbf{Dataset construction and closed-loop evaluation pipeline.}
    \textbf{Top (A):} Given a task configuration specifying the target, shot mode,
    spatial/framing elements, and scene, a task-conditioned expert controller
    generates target-relative reference trajectories and executes them via feedback
    control in AirSim. Synchronized RGB frames, UAV/gimbal states, and executed
    actions form the 9,840-episode dataset. \textbf{Bottom (B):} Evaluated policies
    predict action chunks from language instructions, RGB history, and proprioception
    for closed-loop execution. Performance is measured across mobility capability
    (MTC, NTP, EDP) and cinematography capability (G-TVR, G-FSR, G-TVR-all).}
    \label{fig:construction_evaluation}
\end{figure*}

\noindent\textbf{Task Instance Generation.}
We instantiate each cinematography semantic across varied subjects, scenes, and initial conditions. For every task instance, compatible subjects and feasible flight regions are selected while initial positions, viewing angles, and motion parameters are randomized. In total, we collect 9,840 filming episodes across the ten environments.

\noindent\textbf{Expert Demonstration Collection.}
We design a task-conditioned expert controller to generate reference demonstrations (Fig.~\ref{fig:construction_evaluation}A). Given a task specification, the controller computes a target-relative reference trajectory and desired camera orientation. A feedback tracking controller then follows this reference using simulated UAV states in AirSim. Viewing directions are defined relative to calibrated forward headings for static subjects, and relative to validated motion headings for moving subjects. During flight, we record onboard RGB images, UAV and gimbal states, executed actions, and task metadata, capturing actual closed-loop flight dynamics rather than idealized reference paths.
% Keep controller implementation details in the supplement.
{Reference generation and camera control are detailed in Appendix~\ref{app:expert_control}.}

\noindent\textbf{Trajectory Diversification.}
To broaden state-action coverage beyond nominal paths, we incorporate controlled trajectory variations. \emph{Clean} episodes sample diverse initial positions and filming parameters. \emph{Medium-variation} episodes introduce smooth single- or dual-axis deviations around the nominal reference while preserving the target semantics. \emph{Strong-recovery} episodes inject larger external disturbances and record the expert's corrective recovery maneuvers. Finite-motion tasks include clean, medium-variation, and strong-recovery episodes. Continuous-following tasks use clean and medium-variation episodes with diverse initial conditions, target-acquisition states, and smooth trajectory variations.

\noindent\textbf{Quality Assurance.}
Before full-scale collection, subject compatibility, motion paths, and flight feasibility are validated through pilot runs for every semantic--subject pair. All recorded episodes undergo automated kinematic consistency checks and manual video inspection to filter out wrong motion directions, collisions, or framing failures, and invalid recordings are recollected while preserving the intended cinematography semantics, with compatible subjects reselected when necessary.

\subsection{Dataset Structure and Statistics}
\label{sec:dataset_statistics}

The resulting dataset comprises 9,840 filming episodes across ten maps and 41 concrete cinematography semantics (Fig.~\ref{fig:dataset_overview}D). Of these, 6,240 episodes feature static subjects and 3,600 feature moving subjects; 7,200 are {clean or medium-variation demonstrations} and 2,640 are strong-recovery episodes. By execution regime, 1,920 episodes represent continuous filming and 7,920 represent finite-motion tasks.

Each episode stores the natural-language instruction, timestamped RGB observations, UAV and gimbal states, executed pose increments, and scene metadata. For policy training, each supervision sample consists of three RGB frames sampled at $0.2$-second intervals, a 32-dimensional padded proprioceptive vector, and a future 50-step action chunk ($10$-second horizon at $0.2$ seconds per step). Action labels are computed from actual recorded pose increments rather than raw velocity commands, with task-specific masks indicating active control dimensions.

We split the dataset into 8,200 training episodes and 1,640 validation episodes. Within each map, every pairing of a concrete cinematography semantic with a compatible subject is split into ten training and two validation episodes. The validation split comprises clean and medium-variation episodes, while all strong-recovery episodes are placed in the training set. Our timestamp-aligned export yields 721,720 observation--action samples for training and validation, while full episodes serve as the unit of closed-loop evaluation. {Export accounting, state/action definitions, and stable-tail handling are detailed in Appendix~\ref{app:data_accounting}--\ref{app:data_tail}.}

% !TeX root = ../main.tex

\section{Experiments}
\label{sec:experiments}

\subsection{Evaluation Metrics}
\label{sec:experiment_metric_overview}

Aerial cinematography requires a UAV to translate natural-language instructions into purposeful camera motions while maintaining a desired visual relationship with the subject. Building on the task space in Sec.~\ref{sec:task_taxonomy}, we evaluate five core requirements: \emph{motion correctness}, \emph{task progress}, \emph{subject retention}, \emph{image framing}, and \emph{extraneous motion}. We report six complementary metrics: MTC, NTP, G-TVR, G-TVR-all, G-FSR, and EDP. Rather than requiring pointwise agreement with a single expert trajectory, our evaluation measures whether the requested motion and image-space constraints are satisfied, allowing multiple valid executions of the same instruction.

% !TeX root = ../main.tex
\begin{table}[t]
    \centering
    \caption{Evaluation metrics and their corresponding filming requirements.
    Scores are reported on a $0$--$100$ scale;
    $\uparrow$ indicates higher is better and $\downarrow$ lower is better.
    {Definitions and aggregation are given in Appendix~\ref{app:metric_motion}--\ref{app:metric_aggregation}, with decision thresholds in Appendix~\ref{app:metric_parameters}.}}
    \label{tab:metrics_overview}
    \begingroup
    \small
    \setlength{\tabcolsep}{3pt}
    \renewcommand{\arraystretch}{1.15}
    \begin{tabular}{@{}p{0.26\columnwidth}p{0.22\columnwidth}p{\dimexpr0.52\columnwidth-4\tabcolsep\relax}@{}}
        \toprule
        \textbf{Metric} & \textbf{Requirement} & \textbf{Evaluated aspect} \\
        \midrule
        MTC $\uparrow$ & Motion correctness & Correct primary motion and required joint trends. \\
        NTP $\uparrow$ & Task progress & Finite-motion completion or continuous-following maintenance. \\
        G-TVR $\uparrow$ & Subject retention & Full-task time fraction with the subject reference point in frame. \\
        G-TVR-all $\uparrow$ & Subject retention & Complete execution with uninterrupted subject retention. \\
        G-FSR $\uparrow$ & Image framing & Full-task time fraction with the reference point in the requested region. \\
        EDP $\downarrow$ & Extraneous motion & Unintended height or distance drift and terminal overshoot. \\
        \bottomrule
    \end{tabular}
    \endgroup
\end{table}

\subsection{Experimental Setup}
\label{sec:experimental_setup}

\noindent\textbf{Test set.}
The fixed evaluation benchmark comprises 2,460 test cases across ten maps and 41 concrete cinematography semantics. Each semantic is evaluated on two maps with three subjects per map and ten repetitions per subject, yielding 60 cases per semantic. Of the 2,460 cases, 1,640 use semantic--subject pairs seen during training and 820 evaluate unseen pairs.
% Identify the fixed case-construction subsection, not a different test set.
{Case allocation and repetitions are specified in Appendix~\ref{app:test_construction}.}

\noindent\textbf{Models and interfaces.}
We benchmark nine representative policies spanning three model families: pretrained VLAs (PI0{~\cite{BlackK-RSS-25}}, PI0.5{~\cite{intelligence2025pi05visionlanguageactionmodelopenworld}}, SmolVLA{~\cite{shukor2025smolvlavisionlanguageactionmodelaffordable}}, WALL-OSS 0.5{~\cite{yu2026walloss05technicalreport}}, and OpenVLA-OFT{~\cite{kim2024openvlaopensourcevisionlanguageactionmodel,kim2025finetuningvisionlanguageactionmodelsoptimizing}}), world action model (WAM)-derived policies (Cosmos Policy{~\cite{kim2026cosmospolicyfinetuningvideo}} and DreamZero (5B task adaptation){~\cite{ye2026worldactionmodelszeroshot}}), and language-conditioned action policies (ACT{~\cite{zhao2023learningfinegrainedbimanualmanipulation}} and Diffusion Policy{~\cite{chi2024diffusionpolicyvisuomotorpolicy}}). All models are adapted on the common dataset split in Sec.~\ref{sec:benchmark}, with pretrained policies fine-tuned on the photography data and monitored on the validation split. At each decision step, policies receive an English instruction, three RGB frames at the current timestamp and $0.2$ and $0.4$ seconds earlier (H3), and UAV proprioceptive states. Each policy outputs a sequence of six-dimensional pose increments at $0.2$-second intervals, covering body-frame translation, UAV yaw, and relative gimbal pitch and yaw changes, with task-dependent masks excluding inactive dimensions. {Model interfaces, adaptation details, and evaluated training settings are summarized in Appendix~\ref{app:model_contract}--\ref{app:model_training}.}

% B.1 contains initialization, controller, budgets, and early-stop rules.
% Keep the short protocol together instead of leaving a line under the floats.
\begin{samepage}
{Evaluation pauses simulation during inference, then executes ten predicted steps ($2$ seconds) through a shared tracker before updating observations. Initialization, tracking parameters, task budgets, and early-stop rules are specified in Appendix~\ref{app:execution_limitations}.}
\par\end{samepage}

\subsection{Overall Results}
\label{sec:experiment_results}

% !TeX root = ../main.tex
% Generated by evaluation_analysis/paper_layout_cvpr2027_20261114_v1/render_tables.py
% Revision: Move the complete five-family comparison into the main text as a wide table.
\begin{table*}[!t]
    \centering
    \caption{Complete six-metric results for the five task families, macro-averaged over applicable semantics within each family on a 0--100 scale. Lower is better for EDP. Bold denotes the best value within each family and metric, including ties.}
    \label{tab:task_family_results}
    \begingroup
    \footnotesize
    \setlength{\tabcolsep}{0.8pt}
    \renewcommand{\arraystretch}{1.16}
    \resizebox{\linewidth}{!}{%
    \begin{tabular}{@{}l*{30}{r}@{}}
        \toprule
        \multirow{2}{*}{Model}
        & \multicolumn{6}{c}{Follow (F)}
        & \multicolumn{6}{c}{Orbit (O)}
        & \multicolumn{6}{c}{Elevation (E)}
        & \multicolumn{6}{c}{Fly-over (V)}
        & \multicolumn{6}{c}{Composite (C)} \\
        \cmidrule(lr){2-7}\cmidrule(lr){8-13}\cmidrule(lr){14-19}\cmidrule(lr){20-25}\cmidrule(lr){26-31}
        & MTC & NTP & G-TVR & \shortstack{G-TVR\\all} & G-FSR & EDP & MTC & NTP & G-TVR & \shortstack{G-TVR\\all} & G-FSR & EDP & MTC & NTP & G-TVR & \shortstack{G-TVR\\all} & G-FSR & EDP & MTC & NTP & G-TVR & \shortstack{G-TVR\\all} & G-FSR & EDP & MTC & NTP & G-TVR & \shortstack{G-TVR\\all} & G-FSR & EDP \\
        \midrule
        PI0.5 & 81.33 & 53.11 & 90.47 & 72.50 & 48.72 & 36.17 & 88.56 & 84.64 & 96.14 & 88.89 & 86.07 & 25.80 & 96.39 & 94.86 & 98.94 & 97.50 & 80.01 & 1.52 & \textbf{100.00} & 81.15 & 96.60 & 82.50 & 78.58 & 22.31 & 71.39 & 78.68 & 95.60 & 85.28 & 65.36 & 31.02 \\
        PI0 & 79.83 & 54.50 & \textbf{91.64} & 75.67 & 45.04 & 31.19 & \textbf{90.33} & 85.06 & 96.29 & 88.44 & 85.65 & 23.94 & 94.44 & 93.68 & 98.57 & 97.22 & 77.32 & 1.76 & \textbf{100.00} & \textbf{81.25} & 96.77 & 75.00 & 80.92 & 22.68 & \textbf{75.00} & \textbf{79.72} & 92.49 & 74.72 & 65.17 & 24.65 \\
        OpenVLA-OFT & 18.33 & 19.78 & 69.58 & 35.33 & 30.75 & 48.41 & 79.67 & 67.36 & 74.98 & 44.67 & 45.62 & 41.22 & 81.94 & 88.54 & \textbf{100.00} & \textbf{100.00} & 78.16 & 2.04 & 76.25 & 74.06 & 93.93 & 60.83 & 64.73 & 48.69 & 43.61 & 61.67 & 77.28 & 55.56 & 42.87 & 23.21 \\
        SmolVLA & 81.00 & 57.75 & 88.22 & 66.83 & 50.57 & 28.71 & 88.22 & 78.92 & 95.53 & 88.33 & 83.24 & 29.47 & \textbf{98.33} & 94.51 & 99.50 & 98.89 & 81.68 & 1.62 & 97.92 & 75.83 & 92.84 & 70.42 & 70.08 & 39.97 & 73.61 & 76.18 & 93.66 & 80.56 & 61.58 & \textbf{20.09} \\
        ACT & 77.33 & 44.44 & 89.10 & 75.50 & 45.22 & 44.34 & 89.22 & \textbf{86.50} & 95.94 & 89.56 & 85.42 & 26.54 & 91.94 & 94.10 & 99.68 & 97.22 & 81.17 & 1.47 & \textbf{100.00} & 79.38 & 89.81 & 65.42 & 69.76 & 28.85 & 67.22 & 77.57 & 93.12 & 76.67 & 63.73 & 23.95 \\
        \shortstack[l]{Diffusion\\Policy} & \textbf{83.83} & 59.05 & 91.57 & 77.83 & 51.79 & 26.04 & 86.33 & 83.06 & \textbf{96.95} & \textbf{92.44} & \textbf{88.80} & \textbf{19.32} & 98.06 & \textbf{95.76} & 99.58 & 98.61 & 79.94 & 1.82 & 99.17 & 79.27 & 97.15 & 89.17 & \textbf{83.39} & 23.55 & 69.72 & 79.31 & \textbf{97.49} & \textbf{91.94} & \textbf{74.31} & 26.66 \\
        \shortstack[l]{Cosmos\\Policy} & 57.00 & 39.60 & 74.88 & 41.50 & 29.37 & 37.91 & 86.44 & 73.50 & 90.37 & 79.22 & 63.21 & 37.27 & 96.39 & 89.38 & \textbf{100.00} & \textbf{100.00} & 73.65 & 1.60 & 96.25 & 37.40 & 84.75 & 40.42 & 43.63 & 60.57 & 71.11 & 66.53 & 83.58 & 57.78 & 47.73 & 36.41 \\
        WALL-OSS 0.5 & 83.67 & \textbf{63.94} & 90.47 & 70.50 & \textbf{64.53} & \textbf{21.09} & 83.78 & 83.47 & 94.10 & 88.89 & 80.14 & 24.56 & 98.06 & 94.93 & 99.26 & 98.61 & \textbf{83.28} & \textbf{0.85} & \textbf{100.00} & 80.42 & \textbf{98.09} & \textbf{89.58} & 75.91 & \textbf{21.10} & 73.33 & 71.18 & 91.57 & 78.61 & 59.22 & 26.86 \\
        DreamZero & 73.67 & 54.52 & 90.68 & \textbf{78.00} & 51.36 & 29.70 & 84.33 & 66.81 & 95.40 & 87.33 & 71.93 & 32.86 & 87.22 & 78.06 & 99.93 & 99.72 & 70.65 & 4.85 & 92.92 & 69.17 & 92.35 & 65.00 & 63.56 & 30.02 & 68.06 & 64.79 & 90.92 & 71.11 & 52.09 & 34.68 \\
        \bottomrule
    \end{tabular}
    }
    \endgroup
\end{table*}

Table~\ref{tab:task_family_results} provides the full breakdown across the five task families, while Table~\ref{tab:overall_model_results} reports the overall performance of all nine models across the 2,460 test cases. Diffusion Policy achieves the strongest overall balance across subject retention, framing, and extraneous-motion control, leading G-TVR (96.12), G-TVR-all (89.39), G-FSR (75.83), and EDP (19.54). PI0 achieves the highest motion trend correctness (87.07), while WALL-OSS 0.5 leads normalized task progress (78.29). OpenVLA-OFT ranks lowest across all six metrics. These results show that motion execution, task advancement, and visual composition favor different model designs.

% !TeX root = ../main.tex
% Generated by evaluation_analysis/paper_layout_cvpr2027_20261114_v1/render_tables.py
% Revision: Add DreamZero and keep the overall comparison within one column.
% Revision: Bold column-wise best scores; minimize EDP and maximize other metrics.
\begin{table}[!tbp]
    \centering
    \caption{Complete overall results for nine models on the same 2,460 cases. Scores are macro-averaged over applicable concrete semantics and reported on a 0--100 scale; lower is better for EDP and higher for the other metrics. Bold denotes the best value in each column, including ties.}
    \label{tab:overall_model_results}
    \begingroup
    \footnotesize
    \setlength{\tabcolsep}{2pt}
    \renewcommand{\arraystretch}{1.05}
    \resizebox{\linewidth}{!}{%
    \begin{tabular}{@{}lrrrrrr@{}}
        \toprule
        Model & \shortstack{MTC\\$\uparrow$} & \shortstack{NTP\\$\uparrow$} & \shortstack{G-TVR\\$\uparrow$} & \shortstack{G-TVR-all\\$\uparrow$} & \shortstack{G-FSR\\$\uparrow$} & \shortstack{EDP\\$\downarrow$} \\
        \midrule
        PI0.5 & 86.54 & 77.23 & 95.13 & 85.00 & 72.31 & 24.90 \\
        PI0 & \textbf{87.07} & 77.71 & 94.98 & 83.29 & 71.07 & 22.33 \\
        OpenVLA-OFT & 59.43 & 58.68 & 79.51 & 53.66 & 48.22 & 35.96 \\
        SmolVLA & 86.75 & 75.34 & 93.79 & 81.75 & 70.59 & 25.11 \\
        ACT & 84.55 & 75.35 & 93.81 & 83.01 & 70.29 & 27.22 \\
        Diffusion Policy & 86.26 & 78.14 & \textbf{96.12} & \textbf{89.39} & \textbf{75.83} & \textbf{19.54} \\
        Cosmos Policy & 79.43 & 63.01 & 86.46 & 66.14 & 52.31 & 34.25 \\
        WALL-OSS 0.5 & 85.89 & \textbf{78.29} & 93.99 & 84.39 & 73.32 & 19.90 \\
        DreamZero & 80.61 & 65.39 & 93.96 & 82.32 & 63.01 & 27.64 \\
        \bottomrule
    \end{tabular}
    }
    \endgroup
\end{table}

Across model families, PI0, PI0.5, SmolVLA, and WALL-OSS 0.5 all exceed 85 in MTC and 75 in NTP, showing strong motion execution and task advancement. This strength stems from combining pretrained vision-language representations with continuous action generation, whereas OpenVLA-OFT's lower scores highlight the need to align pretrained representations with the aerial incremental action space. Language-conditioned action policies (ACT and Diffusion Policy) surpass 93 in G-TVR without a pretrained VLA backbone, indicating that learning action chunks directly from cinematography data supports reliable subject retention. Between the two, Diffusion Policy achieves higher G-FSR and lower EDP than ACT, as joint denoising captures temporal dependencies across action sequences to smooth continuous control and reduce extraneous motion. By contrast, WAM-derived policies (Cosmos Policy and DreamZero) show a clear gap between subject retention and task completion: while DreamZero matches leading VLAs in G-TVR, both WAM-derived policies trail the top four VLAs, ACT, and Diffusion Policy in NTP and G-FSR. This gap reflects the difficulty of aligning video-pretrained representations with fine-grained pose control, leaving task advancement and precise framing as their main bottlenecks.

% Move the approved filming-outcome synthesis into the overall results.
{From practical filming requirements, current policies can execute basic camera motions. For example, in ascending, descending, and fixed-radius \emph{Orbit} shots, stronger policies advance the requested altitude or azimuth changes while keeping the subject in frame. Different models also show distinct strengths across shot types: Diffusion Policy achieves better framing retention on \emph{Orbit} tasks, whereas WALL-OSS 0.5 leads in progress and framing metrics on \emph{Follow} tasks. However, moving in the right direction does not ensure sustained cinematic intent. Rear or side \emph{Follow} requires the UAV to maintain a consistent relative viewpoint while the subject moves, and compositional motions such as rising while pulling away demand simultaneous coordination across multiple spatial axes. Under these demands, current policies struggle to balance trajectory advancement with target-relative visual relationships. Overall, existing models can perform basic filming actions, yet fail to consistently leverage visual feedback to unify subject motion, relative geometry, and image framing within continuous control.}

\subsection{Analysis by Task Attributes}
\label{sec:task_attribute_analysis}

% Keep one top float per column instead of stacking both comparison tables.
\setcounter{topnumber}{1}

The fixed evaluation benchmark spans diverse subject-motion states, execution regimes, spatial motions, and framing constraints. We analyze policy capabilities along three key dimensions: motion execution versus visual framing, static versus moving subjects, and multi-constraint coordination. {Complete six-metric comparisons across subject-motion and requirement groups appear in Appendix~\ref{app:full_results} (Tables~\ref{tab:subject_motion_all_metrics}--\ref{tab:semantic_requirement_all_metrics}); semantic group membership is listed in Appendix~\ref{app:semantic_groups}.}

\noindent\textbf{Motion execution and visual performance.}
Evaluating across all 41 semantics reveals a clear decoupling between 3D flight execution and 2D image composition. Most policies achieve high MTC scores, which aligns with our trajectory analysis and manual video inspection confirming accurate primary motions. In contrast, G-FSR lags behind G-TVR by 20.29 to 34.15 points across every model. While current policies execute basic camera trajectories and keep subjects broadly within the field of view, sustaining specified image framing throughout motion remains a shared limitation.

\noindent\textbf{Static and moving subjects.}
The fixed test set is grouped by subject motion state into static-subject tasks (26 semantics; 1,560 cases) and moving-subject tasks (15 semantics; 900 cases), enabling analysis of how policies coordinate subject tracking with cinematographic motion. As shown in Table~\ref{tab:subject_motion_comparison}, all nine models experience substantial drops in both NTP and G-FSR on moving subjects compared to static subjects. Advancing cinematographic trajectories while maintaining framing around dynamic targets remains a common challenge across all evaluated architectures.

% Declare comparisons later so they do not float onto the previous results page.
% !TeX root = ../main.tex
\begin{table}[t]
    \centering
    \caption{Comparison between static-subject (26 semantics) and moving-subject (15 semantics) tasks on the fixed test set. Scores are on a 0--100 scale, where higher is better. Bold denotes the best score in each column.}
    \label{tab:subject_motion_comparison}
    \begingroup
    \small
    \setlength{\tabcolsep}{3pt}
    \renewcommand{\arraystretch}{1.12}
    \begin{tabular}{@{}lrrrr@{}}
        \toprule
        \multirow{2}{*}{Model} & \multicolumn{2}{c}{NTP $\uparrow$} & \multicolumn{2}{c}{G-FSR $\uparrow$} \\
        \cmidrule(lr){2-3}\cmidrule(lr){4-5}
        & Static & Moving & Static & Moving \\
        \midrule
        PI0.5 & 86.39 & 61.35 & 84.19 & 51.71 \\
        PI0 & 86.30 & 62.83 & 82.71 & 50.89 \\
        OpenVLA-OFT & 71.14 & 37.08 & 55.21 & 36.09 \\
        SmolVLA & 82.47 & 62.97 & 81.75 & 51.24 \\
        ACT & 85.91 & 57.04 & 81.82 & 50.31 \\
        Diffusion Policy & 85.10 & \textbf{66.09} & \textbf{85.90} & 58.36 \\
        Cosmos Policy & 72.92 & 45.84 & 63.71 & 32.54 \\
        WALL-OSS 0.5 & \textbf{86.97} & 63.24 & 81.13 & \textbf{59.77} \\
        DreamZero & 69.33 & 58.57 & 71.18 & 48.84 \\
        \bottomrule
    \end{tabular}
    \endgroup
\end{table}

% !TeX root = ../main.tex
\begin{table}[t]
    \centering
    \caption{Comparison between single-requirement (20 semantics) and joint-requirement (21 semantics) tasks on the fixed test set. Scores are on a 0--100 scale, where higher is better. Bold denotes the best score in each column.}
    \label{tab:semantic_requirement_comparison}
    \begingroup
    \small
    \setlength{\tabcolsep}{3pt}
    \renewcommand{\arraystretch}{1.12}
    \begin{tabular}{@{}lrrrr@{}}
        \toprule
        \multirow{2}{*}{Model} & \multicolumn{2}{c}{MTC $\uparrow$} & \multicolumn{2}{c}{G-FSR $\uparrow$} \\
        \cmidrule(lr){2-3}\cmidrule(lr){4-5}
        & Single & Joint & Single & Joint \\
        \midrule
        PI0.5 & 93.92 & 79.52 & 78.89 & 66.05 \\
        PI0 & 93.50 & 80.95 & 77.16 & 65.27 \\
        OpenVLA-OFT & 68.33 & 50.95 & 52.95 & 43.71 \\
        SmolVLA & 92.50 & \textbf{81.27} & 76.98 & 64.51 \\
        ACT & 92.58 & 76.90 & 76.74 & 64.15 \\
        Diffusion Policy & 92.75 & 80.08 & 78.69 & \textbf{73.10} \\
        Cosmos Policy & 86.83 & 72.38 & 58.93 & 46.00 \\
        WALL-OSS 0.5 & \textbf{94.08} & 78.10 & \textbf{79.53} & 67.40 \\
        DreamZero & 87.25 & 74.29 & 69.55 & 56.78 \\
        \bottomrule
    \end{tabular}
    \endgroup
\end{table}

% Keep the short comparison paragraph with its heading.
\begin{samepage}
\noindent\textbf{Joint requirements and multi-constraint coordination.}
We further compare 20 single-requirement semantics (1,200 cases per model), which specify one primary motion or viewpoint, against 21 joint-requirement semantics (1,260 cases per model), which combine multiple spatial motions, co-motion, or thirds framing constraints. As shown in Table~\ref{tab:semantic_requirement_comparison}, introducing joint requirements reduces MTC by 11.23 to 17.38 points and G-FSR by 5.58 to 12.94 points across all nine models. This consistent decline confirms that coordinating multiple simultaneous filming constraints is a shared bottleneck across model families.
\par\end{samepage}

% Keep the heading with its opening lines without moving the whole conclusion.
\Needspace{3\baselineskip}

\section{Conclusion}
\label{sec:conclusion}

In this work, we introduce \textbf{UAV-CinePilot}, a closed-loop benchmark and dataset for language-guided UAV cinematography. The benchmark establishes 41 concrete cinematography semantics across ten simulation environments, comprising 9,840 filming episodes. It evaluates policy performance using six complementary metrics that jointly measure motion execution and visual composition. Systematic evaluations of nine representative policies show that current models can translate basic instructions into executable camera motions, yet struggle with moving subjects, precise framing, and multi-constraint coordination. \textbf{UAV-CinePilot} provides a standardized platform for benchmarking aerial cinematography policies and lays the groundwork for future research into language grounding, temporal visual perception, and continuous control coordination.

% Keep main-text floats ahead of the statement and bibliography.
\FloatBarrier
% !TeX root = ../main.tex
\section*{LLM Usage Statement}

The authors led this research and are responsible for the methodology,
experiments, conclusions, and final manuscript. OpenAI Codex and OpenAI
image tools provided assistance with implementation, analysis, language
editing, and illustrative asset preparation.

{\small
\bibliographystyle{ieeenat_fullname}
\bibliography{main}
}

\clearpage
\appendix
\numberwithin{equation}{section}
\setcounter{figure}{0}
\renewcommand{\theHfigure}{supp.\arabic{figure}}
% Appendix-only figure formatting, loaded after the main paper.
\renewcommand{\thefigure}{S\arabic{figure}}
\captionsetup{font=small,labelfont=bf}
\setlength{\textfloatsep}{12pt plus 2pt minus 2pt}
\setlength{\floatsep}{10pt plus 2pt minus 2pt}
\setlength{\intextsep}{12pt plus 2pt minus 2pt}
\setlength{\dbltextfloatsep}{12pt plus 2pt minus 2pt}
\setlength{\dblfloatsep}{10pt plus 2pt minus 2pt}
\renewcommand{\dbltopfraction}{0.85}
\renewcommand{\dblfloatpagefraction}{0.70}
\makeatletter
\setlength{\@dblfptop}{0pt}
\setlength{\@dblfpsep}{12pt plus 2pt}
\makeatother
\newenvironment{suppfigure}[1][\textwidth]{%
    \begin{figure*}[!tp]\centering
    \begin{minipage}{#1}\centering
}{%
    \end{minipage}
    \end{figure*}}

\renewcommand{\dblfloatpagefraction}{0.90}
\raggedbottom
% !TeX root = ../../main.tex
% Detailed metric definitions for Appendix A.

\section{Evaluation Metrics}
\label{sec:evaluation_metrics}

We define the six metrics introduced in Sec.~4.1 of the main paper and specify their scoring criteria for finite motions and continuous following.

\subsection{Motion Correctness and Task Progress}
\label{app:metric_motion}

\noindent\textbf{Motion trend correctness (MTC).}
MTC evaluates whether the policy executes the requested primary motion (e.g., orbiting clockwise rather than counterclockwise). Let $b_n \in \{0,1\}$ denote the binary trend outcome for task $n$:
\begin{equation}
    \mathrm{MTC}_n=b_n.
    \label{eq:metric_mtc}
\end{equation}
MTC considers both signed net displacement and the direction of travel over the trajectory. All specified axes must pass, with concurrent progress where the instruction requires simultaneous changes. For continuous following, MTC checks subject-relative bearing and speed.

\noindent\textbf{Normalized task progress (NTP).}
To quantify execution completeness, we define NTP separately for finite camera motions and continuous following:
\begin{equation}
    \mathrm{NTP}_n=
    \begin{cases}
        K_n/4, & \text{finite motion},\\
        \displaystyle\frac{\sum_j L_jc_{n,j}}{T_n},
        & \text{continuous following}.
    \end{cases}
    \label{eq:metric_ntp}
\end{equation}
For finite motions, $K_n \in \{0,1,2,3,4\}$ counts the stages attained in order: early, middle, late, and terminal confirmation. The first three stages follow spatial milestones, not elapsed time. Revisiting a completed stage adds no credit. Terminal confirmation also requires completing the task budget and holding a stable terminal configuration at its end.

For continuous following, $T_n$ is the full task budget, $L_j$ is the scheduled duration of window $j$, and $c_{n,j} \in \{0,1\}$ indicates whether that window passes. A window passes if bearing, distance, relative height, and relative speed jointly satisfy their criteria for the required fraction of its scheduled duration.

MTC and NTP capture complementary aspects of execution. A partial orbit can pass MTC without earning full NTP, while reaching a milestone does not establish a consistent motion trend.

\subsection{Subject Retention and Framing}
\label{app:metric_framing}

Correct 3D motion does not guarantee correct framing. Let $g_{n,i} \in \{0,1\}$ indicate whether the subject reference point projects inside the camera field of view at evaluation step $i$, $w_{n,i}$ be its supported duration, $\mathbf z_{n,i}$ its normalized image coordinates, and $\Omega_n$ the target framing region. Defining $f_{n,i}=g_{n,i}\mathbf{1}[\mathbf z_{n,i}\in\Omega_n]$, we compute:
\begin{equation}
    \begin{aligned}
        \mathrm{G\!\text{-}\!TVR}_n
        &=\frac{\max(0,\sum_i w_{n,i}g_{n,i}-R_{g,n})}{T_n},\\
        \mathrm{G\!\text{-}\!TVR\!\text{-}\!all}_n
        &=\mathbf{1}[\text{complete, always in frame}],\\
        \mathrm{G\!\text{-}\!FSR}_n
        &=\frac{\max(0,\sum_i w_{n,i}f_{n,i}-R_{f,n})}{T_n}.
    \end{aligned}
    \label{eq:metric_retention_framing}
\end{equation}
Here, $\mathbf{1}[\cdot]$ is the indicator function. Pass durations are first accumulated on the $0.2$-s grid, which can miss brief failures recorded in the finer control logs. $R_{g,n}$ and $R_{f,n}$ therefore subtract the overlap between grid-passing intervals and raw-log evidence of an out-of-frame or framing failure, respectively.

\textbf{Geometric target-in-view ratio (G-TVR)} measures the fraction of the full budget during which the subject reference point lies within the field of view.
\textbf{Full-task in-view indicator (G-TVR-all)} equals one only if execution reaches the full budget with no geometric out-of-frame event.
\textbf{Geometric framing satisfaction ratio (G-FSR)} measures the fraction of the budget with the reference point inside $\Omega_n$. Standard shots use a central region; F7 and E5 use the specified thirds region.

\begin{figure*}[!tp]
    \centering
    \includegraphics[width=0.98\linewidth]{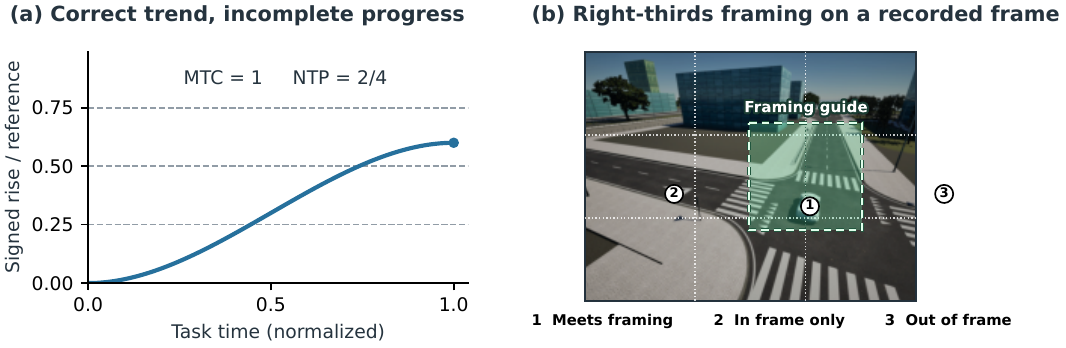}
    \caption{\textbf{Illustrative metric decisions.} (a) The example rise follows the requested direction and reaches $60\%$ of its reference amplitude, passing MTC but only two NTP milestones. (b) An onboard F7 training frame with a shaded right-thirds guide. The guide is illustrative; Appendix~\ref{app:metric_parameters} gives the scoring bounds. Marker~1 is placed on the visible vehicle; markers~2 and~3 show hypothetical placements. G-TVR credits time in states~1 and~2; G-FSR credits only state~1. An out-of-frame event such as state~3 makes G-TVR-all zero. Neither panel contributes to the reported scores.}\label{fig:metric_examples}
\end{figure*}

G-TVR and G-FSR both use the full budget $T_n$: out-of-frame intervals receive no credit, but do not erase credit from other intervals. The ``G'' prefix denotes geometric projection, not pixel-level visibility. Figure~\ref{fig:metric_examples} illustrates the distinction.

\subsection{Extraneous Motion and Aggregation}
\label{app:metric_aggregation}

\noindent\textbf{Excessive displacement penalty (EDP).}
Even when the primary motion follows the instruction, the UAV can drift in altitude or distance or overshoot the terminal region. EDP measures sustained extraneous motion over executed windows:
\begin{equation}
    \mathrm{EDP}_n=
    \frac{\sum_j\ell_{n,j}e_{n,j}}{\sum_j\ell_{n,j}}.
    \label{eq:metric_edp}
\end{equation}
Here, $\ell_{n,j}$ is the evaluable duration of window $j$, and $e_{n,j} \in \{0,1\}$ indicates whether it is flagged. A window is flagged when both the abnormal time fraction and the continuous abnormal duration meet their thresholds. The geometric criteria distinguish normal corrections from drift or overshoot.

Lower EDP indicates less sustained extraneous motion. EDP uses only evaluable executed windows, so we interpret it alongside NTP and G-TVR. Semantics without fixed holding axes that are excluded from EDP receive no score rather than zero.

We score each test episode, including safety-stopped runs, on a $0$--$100$ scale. We first average cases within each concrete semantic, then macro-average applicable semantics with equal weight, both overall and within each task family. Metrics normalized by $T_n$ retain the full denominator after an early stop. We report the six metrics separately.

% Supply the frozen scoring criteria without changing any reported result.
\begingroup

\subsection{Frozen Decision Criteria}
\label{app:metric_parameters}

\noindent\textbf{Geometry and time.}
Let $\mathbf r(t)=\mathbf p_{\mathrm d}(t)-\mathbf p_{\mathrm s}(t)$ denote the measured UAV--subject displacement in world NED. The horizontal radius is $r=\|\mathbf r_{xy}\|$, and relative height is $h=-r_z$. Static-subject bearings use the subject's calibrated forward axis. Moving-subject bearings use the locked motion axis when specified; otherwise the heading follows measured velocity at speeds of at least $0.3\,$m/s, with a $0.3$-second smoothing constant. Orbit trials accumulate unwrapped azimuth in a fixed world-horizontal reference initialized from the case heading, so subject turns alone do not count as orbit progress. Viewed from above, clockwise is positive. Subject-relative front, right, rear, and left correspond to $0^\circ$, $90^\circ$, $180^\circ$, and $-90^\circ$.

The nominal $0.05$-second control logs are aligned to a $0.2$-second scoring grid. We use non-overlapping $2$-second windows and retain the scheduled duration of any final partial window. Grid weights count only supported, executed time within the task budget. We prefer recorded pause-boundary snapshots, use linear interpolation for gaps up to $0.3$ seconds, and velocity-constrained interpolation for longer gaps up to $2$ seconds. Boundary alignment is limited to $0.2$ seconds. Larger unsupported gaps are recording failures. The scorer retains raw evidence of out-of-frame events, framing failures, and crossings.

\noindent\textbf{Finite-motion MTC.}
For each specified axis $a$, $x_a(t)$ is the measured radius, height, unwrapped azimuth, or along-/cross-track coordinate, and $\sigma_a\in\{-1,1\}$ is the required sign. Within each $2$-second window, $\widehat\beta_{a,j}$ is the least-squares slope and $\tau_j$ is the time between the first and last supported samples. The signed fitted change is $d_{a,j}=\sigma_a\widehat\beta_{a,j}\tau_j$. We ignore changes smaller than $\epsilon_a$ ($0.2\,$m for translation and $5^\circ$ for azimuth) when summing directional travel:
\begin{equation}
\begin{aligned}
 P_a&=\sum_{j:|d_{a,j}|\geq\epsilon_a}\max(d_{a,j},0),\\
 N_a&=\sum_{j:|d_{a,j}|\geq\epsilon_a}\max(-d_{a,j},0).
\end{aligned}
\label{eq:metric_directional_travel}
\end{equation}
An axis passes if its signed net change is at least $\epsilon_a$ and $P_a/(P_a+N_a)\geq0.70$; a zero denominator fails. All required axes must pass. The task needs at least $0.4$ seconds of executed evidence and, for moving subjects, at least $0.5\,$m of UAV travel. When simultaneous changes are required, their signed progress must overlap for at least $0.2$ seconds. If the orbit direction is unspecified, its net change determines the sign for the entire trial.

\noindent\textbf{Finite-motion NTP.}
For non-flyover actions, the first three stages require all requested signed changes to reach $25\%$, $50\%$, and $75\%$ of their frozen reference amplitudes in order. Full, half, and quarter orbits use $360^\circ$, $180^\circ$, and $90^\circ$ references, with terminal ranges $[330,390]^\circ$, $[153,207]^\circ$, and $[76.5,103.5]^\circ$, respectively. O8 also requires the terminal bearing to be within $45^\circ$ of the specified cardinal direction. Table~\ref{tab:linear_progress_references} gives the radius and height thresholds.

For flyovers, project the horizontal subject-relative position along and across the required passage to obtain $a$ and $b$. With $L=\max(0.5\,\mathrm m,|a(0)|)$, the stages are approach ($a\geq-L/2$), a negative-to-positive crossing, and exit ($a\geq L/2$). The crossing corridor is $|b|\leq\max(2\,\mathrm m,0.2r(0))$. The terminal region requires $a\in[0.8L,1.6L]$ within that corridor.

The fourth stage requires completing the full budget and holding the terminal configuration over its final $0.6$ seconds. At least $90\%$ of that interval must be supported, all sampled terminal conditions must pass, and there must be no internal gap. Subject-relative UAV speed must be at most $\max(0.8\,\mathrm{m/s},0.3v_{\mathrm s})$, where $v_{\mathrm s}$ is the subject's horizontal speed. Relative velocity accounts for changes in the subject reference heading. Simultaneous spatial changes also require at least $0.2$ seconds of overlapping signed progress. Safety-stopped executions cannot pass the fourth stage.

\noindent\textbf{Continuous following.}
F1, F2, the two F3 sides, the two F4 sides, and the two F7 variants use bearing centers of $180^\circ$, $0^\circ$, $\pm90^\circ$, $\pm135^\circ$, and $180^\circ$, respectively. The tolerance is $60^\circ$, or $45^\circ$ for F4. F3 and F4 require the specified side, and F4 also requires a rear position. Following MTC checks this sector and relative speed at most $\max(1.25\,\mathrm{m/s},0.75v_{\mathrm s})$ for at least $70\%$ of observed execution, with at least $2$ seconds observed.

A following NTP window passes if, for at least $70\%$ of its scheduled duration, the UAV jointly satisfies the sector, $|r-r_0|\leq\max(2\,\mathrm m,0.35r_0)$, $|h-h_0|\leq\max(1.5\,\mathrm m,0.35|h_0|)$, and relative speed at most $\max(1\,\mathrm{m/s},0.5v_{\mathrm s})$. Here $r_0,h_0$ are the initial measured configuration. Early stopping does not shorten the scheduled window.

\noindent\textbf{Image-space regions.}
Normalized image coordinates $(u,v)$ increase rightward and downward. Standard framing requires $|u-0.5|\leq0.15$ and $|v-0.5|\leq0.15$. F7 moves the horizontal center to $1/3$ or $2/3$ with half-width $0.12$; E5 applies the same change on the vertical axis. The other axis keeps its standard tolerance. A reference point is geometrically in view only if it lies in front of the camera and within both field-of-view limits. These metrics do not evaluate pixel-level occlusion. G-TVR-all also checks the original out-of-frame evidence, not just the resampled anchors.

\noindent\textbf{EDP thresholds.}
The holding tolerances around the initial configuration are $\max(2\,\mathrm m,0.30r_0)$ for radius, $\max(1.5\,\mathrm m,0.25|h_0|)$ for height, and $45^\circ$ for fixed bearing. Requested changes, orbit bearing, and flyover radius are not held fixed. Overshoot beyond a requested axis's terminal bound, flyover corridor violations, and departures from the terminal region after confirmed completion also count as abnormal motion.

A window is flagged if abnormal motion occupies at least $35\%$ of its duration and includes a continuous run of at least $\min(0.4\,\mathrm s,\ell_{n,j})$. For finite motions, the end of the first uninterrupted $0.6$ seconds satisfying the terminal conditions after stage three marks the EDP completion boundary. Windows crossing that boundary are split. Unlike the NTP terminal criterion, this boundary does not require an end-of-budget hold. C4 and C5 have no fixed holding axes and are excluded, leaving 39 applicable semantics. EDP is reported independently.
\endgroup

% !TeX root = ../../main.tex
% Full grouped results and execution protocol for Appendix B.
\section{Task-Attribute Results and Test Execution Protocol}
\label{app:full_results}

Tables~\ref{tab:subject_motion_all_metrics} and~\ref{tab:semantic_requirement_all_metrics} report the six metrics for the subject-motion and requirement groups analyzed in main-paper Sec.~4.4. The nine models share the same 2,460 test cases, with metrics aggregated as in Appendix~A. The groups naturally differ in shot geometry and duration and are not single-variable ablations.

The cases in Appendix~\ref{app:visual_examples} illustrate the filming outcomes discussed in Secs.~4.3--4.4: framing drift in C1, following-distance drift in F3, an orbit exceeding its terminal range in O8, and a successful rise and hold in E1.

\begin{table*}[!tp]
\centering
% !TeX root = ../main.tex
% Generated by evaluation_analysis/paper_layout_cvpr2027_20261114_v1/render_tables.py
\par\noindent
\begin{minipage}{\linewidth}
    \centering
    \captionof{table}{All six metrics for static and moving subjects, covering 26 and 15 semantics; lower is better for EDP. Bold indicates the best value in each column, including ties.}
    \label{tab:subject_motion_all_metrics}
    \begingroup
    \footnotesize
    \setlength{\tabcolsep}{3pt}
    \renewcommand{\arraystretch}{1.05}
    \begin{tabular}{@{}l*{12}{r}@{}}
        \toprule
        \multirow{2}{*}{Model} & \multicolumn{2}{c}{MTC $\uparrow$} & \multicolumn{2}{c}{NTP $\uparrow$} & \multicolumn{2}{c}{G-TVR $\uparrow$} & \multicolumn{2}{c}{G-TVR-all $\uparrow$} & \multicolumn{2}{c}{G-FSR $\uparrow$} & \multicolumn{2}{c}{EDP $\downarrow$} \\
        \cmidrule(lr){2-3}\cmidrule(lr){4-5}\cmidrule(lr){6-7}\cmidrule(lr){8-9}\cmidrule(lr){10-11}\cmidrule(lr){12-13}
        & Static & Moving & Static & Moving & Static & Moving & Static & Moving & Static & Moving & Static & Moving \\
        \midrule
        PI0.5 & 87.82 & 84.33 & 86.39 & 61.35 & 97.46 & 91.10 & 92.56 & 71.89 & 84.19 & 51.71 & 17.43 & 36.86 \\
        PI0 & \textbf{88.85} & 84.00 & 86.30 & 62.83 & 96.46 & 92.42 & 88.59 & 74.11 & 82.71 & 50.89 & 16.58 & 31.52 \\
        OpenVLA-OFT & 73.01 & 35.89 & 71.14 & 37.08 & 82.52 & 74.30 & 58.27 & 45.67 & 55.21 & 36.09 & 31.57 & 42.98 \\
        SmolVLA & 88.08 & 84.44 & 82.47 & 62.97 & 96.98 & 88.26 & 90.58 & 66.44 & 81.75 & 51.24 & 22.58 & 29.14 \\
        ACT & 85.45 & 83.00 & 85.91 & 57.04 & 96.38 & 89.35 & 88.01 & 74.33 & 81.82 & 50.31 & 19.40 & 39.74 \\
        Diffusion Policy & 85.64 & \textbf{87.33} & 85.10 & \textbf{66.09} & 97.78 & \textbf{93.24} & \textbf{94.68} & \textbf{80.22} & \textbf{85.90} & 58.36 & \textbf{14.40} & 27.76 \\
        Cosmos Policy & 87.24 & 65.89 & 72.92 & 45.84 & 93.46 & 74.33 & 79.10 & 43.67 & 63.71 & 32.54 & 30.76 & 39.82 \\
        WALL-OSS 0.5 & 87.50 & 83.11 & \textbf{86.97} & 63.24 & \textbf{97.84} & 87.31 & 94.29 & 67.22 & 81.13 & \textbf{59.77} & 16.34 & \textbf{25.60} \\
        DreamZero & 83.78 & 75.11 & 69.33 & 58.57 & 96.87 & 88.91 & 88.85 & 71.00 & 71.18 & 48.84 & 23.69 & 33.96 \\
        \bottomrule
    \end{tabular}
    \endgroup
\end{minipage}
\par

\par\addvspace{10pt}
% !TeX root = ../main.tex
% Generated by evaluation_analysis/paper_layout_cvpr2027_20261114_v1/render_tables.py
\par\noindent
\begin{minipage}{\linewidth}
    \centering
    \captionof{table}{All six metrics for single and joint filming requirements, partitioning all 41 semantics into groups of 20 and 21; lower is better for EDP and inapplicable values are not filled with zero. Bold indicates the best value in each column, including ties.}
    \label{tab:semantic_requirement_all_metrics}
    \begingroup
    \footnotesize
    \setlength{\tabcolsep}{3pt}
    \renewcommand{\arraystretch}{1.05}
    \begin{tabular}{@{}l*{12}{r}@{}}
        \toprule
        \multirow{2}{*}{Model} & \multicolumn{2}{c}{MTC $\uparrow$} & \multicolumn{2}{c}{NTP $\uparrow$} & \multicolumn{2}{c}{G-TVR $\uparrow$} & \multicolumn{2}{c}{G-TVR-all $\uparrow$} & \multicolumn{2}{c}{G-FSR $\uparrow$} & \multicolumn{2}{c}{EDP $\downarrow$} \\
        \cmidrule(lr){2-3}\cmidrule(lr){4-5}\cmidrule(lr){6-7}\cmidrule(lr){8-9}\cmidrule(lr){10-11}\cmidrule(lr){12-13}
        & Single & Joint & Single & Joint & Single & Joint & Single & Joint & Single & Joint & Single & Joint \\
        \midrule
        PI0.5 & 93.92 & 79.52 & 74.57 & 79.77 & 95.80 & 94.49 & 88.58 & 81.59 & 78.89 & 66.05 & 24.28 & 25.55 \\
        PI0 & 93.50 & 80.95 & 75.87 & 79.47 & 95.88 & 94.13 & 87.75 & 79.05 & 77.16 & 65.27 & 21.07 & 23.66 \\
        OpenVLA-OFT & 68.33 & 50.95 & 54.00 & 63.13 & 83.37 & 75.84 & 57.33 & 50.16 & 52.95 & 43.71 & 38.80 & 32.97 \\
        SmolVLA & 92.50 & \textbf{81.27} & 72.89 & 77.67 & 94.40 & 93.21 & 84.67 & 78.97 & 76.98 & 64.51 & 26.11 & 24.05 \\
        ACT & 92.58 & 76.90 & 70.69 & 79.79 & 93.91 & 93.71 & 86.92 & 79.29 & 76.74 & 64.15 & 28.85 & 25.51 \\
        Diffusion Policy & 92.75 & 80.08 & 75.50 & \textbf{80.66} & \textbf{95.99} & \textbf{96.25} & \textbf{90.17} & \textbf{88.65} & 78.69 & \textbf{73.10} & \textbf{18.65} & 20.47 \\
        Cosmos Policy & 86.83 & 72.38 & 62.40 & 63.59 & 88.80 & 84.23 & 70.67 & 61.83 & 58.93 & 46.00 & 35.67 & 32.75 \\
        WALL-OSS 0.5 & \textbf{94.08} & 78.10 & \textbf{78.09} & 78.47 & 94.83 & 93.19 & 86.33 & 82.54 & \textbf{79.53} & 67.40 & 19.82 & \textbf{19.99} \\
        DreamZero & 87.25 & 74.29 & 64.82 & 65.93 & 94.87 & 93.10 & 86.75 & 78.10 & 69.55 & 56.78 & 27.06 & 28.25 \\
        \bottomrule
    \end{tabular}
    \endgroup
\end{minipage}
\par

\end{table*}

\subsection{Test Execution Protocol}
\label{app:execution_limitations}

\noindent\textbf{Initialization.}
For each case, all policies share the shot-specific initialization, subject route, camera parameters, and task budget. The UAV settles at the starting pose. For moving subjects, we collect one second of static history before release. No expert actions are used for warmup.

\noindent\textbf{Shared action execution.}
The simulator pauses for each policy inference, then resumes without repositioning the UAV or subject. A cycle executes up to the first ten $0.2$-s predictions before the next observation. The tracker accumulates displacements from the measured UAV pose, rotating each step into world coordinates using the heading updated by preceding yaw increments. At $20\,$Hz it fits velocity commands to at most five upcoming timed waypoints. During policy execution, subject state is used for scoring and safety, not for waypoint tracking.

\noindent\textbf{Task budgets and stopping.}
For finite-motion cases, the fixed budget is
\begin{equation}
 T_n=2\left\lceil\frac{t_n^{\mathrm{ref}}+2}{2}\right\rceil
 \quad\text{seconds}.
\label{eq:case_budget}
\end{equation}
The reference time $t_n^{\mathrm{ref}}$ comes from the source demonstration's measured completion or endpoint arrival. Finite budgets range from $6$ to $32$ seconds; continuous tasks use $18$ or $24$ seconds. Collision, eight consecutive seconds of geometric subject loss, or a distance or height safety violation ends the test early. Such runs remain in the results, with no visibility, framing, or following-progress credit for unexecuted time. Infrastructure failures are handled separately. The budget excludes inference pauses.

% !TeX root = ../../main.tex
% Data definitions for the standalone supplement.
\begingroup

\section{Data Export and Temporal Supervision}
\label{app:data_protocol}

\subsection{Episode Splits and Export Accounting}
\label{app:data_accounting}

We partition the 9,840 source episodes into 8,200 training and 1,640 validation episodes, with ten training and two validation episodes per collection group. Exported samples inherit their episode split. Table~\ref{tab:export_accounting} reports sample counts after temporal filtering.

\begin{table}[htbp]
    \centering
    \caption{Exported observation--action sample counts. The clean training group includes medium-variation demonstrations. Counts refer to exported individual samples, not source episodes.}
    \label{tab:export_accounting}
    \begin{tabular}{@{}p{0.75\linewidth}r@{}}
        \toprule
        Set & Samples \\
        \midrule
        Clean and medium-variation training & 397,170 \\
        Strong-recovery training & 206,536 \\
        Training total & 603,706 \\
        Validation with fully recorded futures & 113,406 \\
        Separate validation with continued stable tails & 4,608 \\
        \midrule
        Total exported samples & 721,720 \\
        \bottomrule
    \end{tabular}
\end{table}

\subsection{Observations and Incremental Labels}
\label{app:data_labels}

Each sample uses the history $[I_t,I_{t-0.2},I_{t-0.4}]$. We match frames by recorded acquisition timestamps and do not count repeated acquisitions as independent observations. Proprioception comes from the most recent control tick that had closed before the image request. The policy does not receive target position, target velocity, actor identity, event time, or expert progress.

The export admits history frames within $0.075$ seconds of their nominal requested time. It restricts interpolation to poses bracketed by control ticks no more than $0.25$ seconds apart. The aligned own-state incorporates a committed past tick with a conservative age bound of $0.5$ seconds.

The exported action sequence spans 51 temporal boundaries from $t$ to $t+10$ seconds. Bounded interpolation aligns recorded poses to the precise $0.2$-second grid. For each adjacent pair, the exporter rotates world-NED position differences by the UAV yaw at the start of that step into a horizontal heading-aligned frame (forward, right, down). The system recomputes this transform at every step. Translation is measured in metres (downward is positive). UAV yaw increments are wrapped across the angle boundary and expressed in radians.

Gimbal labels use the recorded, applied camera settings relative to the UAV, not physical encoder readings. Pitch increments are downward positive; gimbal yaw is relative to the UAV and does not add a world-yaw increment. The raw proprioceptive vector has 16 active dimensions and 16 zeros. F7 uses the action mask $[1,1,1,0,1,1]$; all other semantics use $[1,1,1,1,1,0]$. Padding and inactive action dimensions do not contribute supervision.

The 32-dimensional proprioceptive vector is indexed as follows:
\begin{center}
\begin{tabular}{@{}cp{0.64\linewidth}l@{}}
\toprule
Indices & Own-state quantity & Units \\
\midrule
1--3 & Measured body-FRD linear velocity & m/s \\
4--6 & Measured body-FRD angular velocity & rad/s \\
7--8 & UAV roll and pitch & rad \\
9--10 & Applied relative camera pitch and yaw & rad \\
11--13 & Most recent commanded body-FRD velocity & m/s \\
14 & Most recent commanded UAV yaw rate & rad/s \\
15--16 & Most recent commanded gimbal yaw and pitch rates & rad/s \\
17--32 & Zero padding & --- \\
\bottomrule
\end{tabular}
\end{center}
The previous command is a single causal control record, not a second action history. The state omits absolute UAV position, absolute yaw, direction codes, and target-speed categories. Camera pitch in the state follows the logged camera convention; the pitch increment label is downward positive.

For adjacent boundaries $t_k,t_{k+1}$, translation and UAV yaw labels are:
\begin{equation}
\begin{aligned}
 \Delta\mathbf p_{b,k}&=R_z(\psi_k)^\top(\mathbf p_{k+1}-\mathbf p_k),\\
 \Delta\psi_{b,k}&=\operatorname{wrap}(\psi_{k+1}-\psi_k).
\end{aligned}
\label{eq:actual_increment_labels}
\end{equation}
Here, $R_z$ is the NED yaw rotation; translation uses a horizontal heading-aligned frame, not a full roll--pitch--yaw rotation. Camera labels are $\Delta\theta_{g,k}=-(\theta^{\rm cam}_{k+1}-\theta^{\rm cam}_k)$ and $\Delta\psi_{g,k}=\operatorname{wrap}(\psi^{\rm rel}_{g,k+1}-\psi^{\rm rel}_{g,k})$, both in radians. During execution, each predicted displacement is rotated using the yaw at the start of that step before accumulation.

Figures~\ref{fig:temporal_scales} and~\ref{fig:coordinate_labels} illustrate the visual history and step-local coordinate transform on a standard C1 recording.

\begin{suppfigure}
\begin{minipage}[t]{0.302\linewidth}\centering\includegraphics[width=\linewidth]{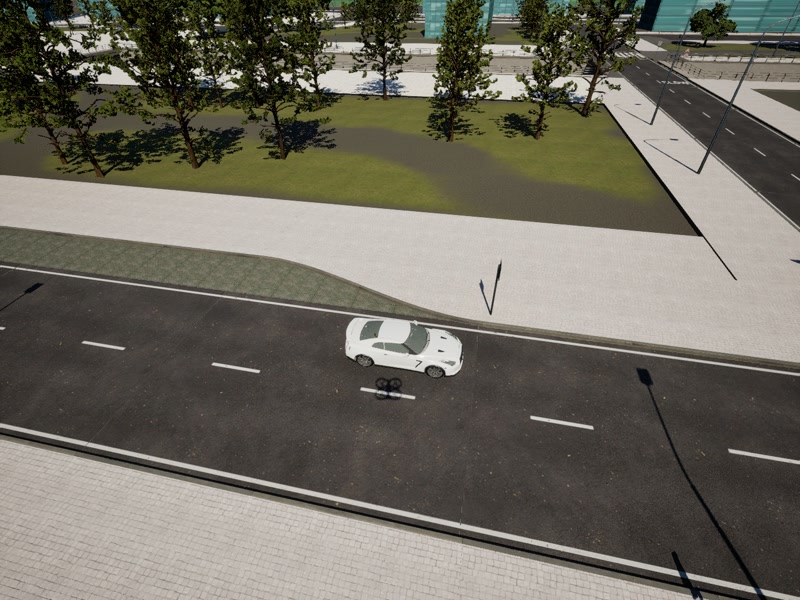}\\{\scriptsize $I_{t-0.400}$: $2.630$ s}\end{minipage}\hfill
\begin{minipage}[t]{0.302\linewidth}\centering\includegraphics[width=\linewidth]{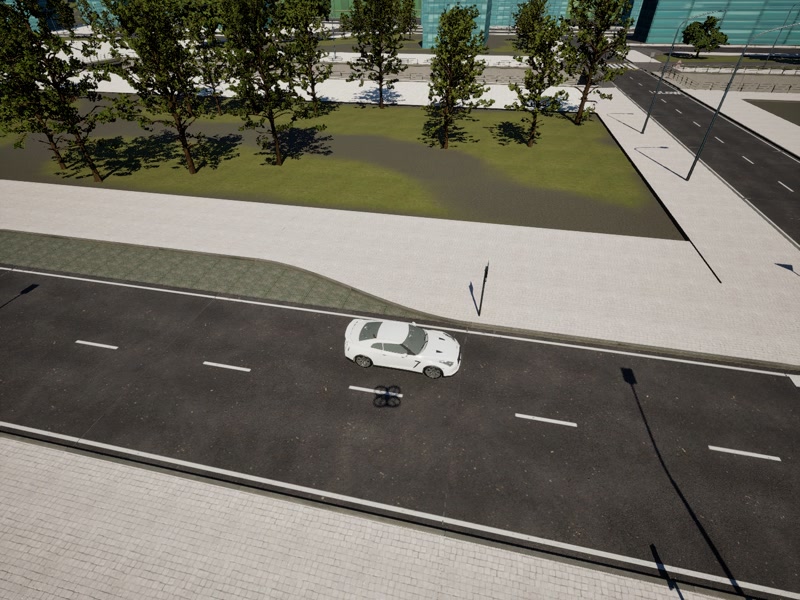}\\{\scriptsize $I_{t-0.197}$: $2.833$ s}\end{minipage}\hfill
\begin{minipage}[t]{0.302\linewidth}\centering\includegraphics[width=\linewidth]{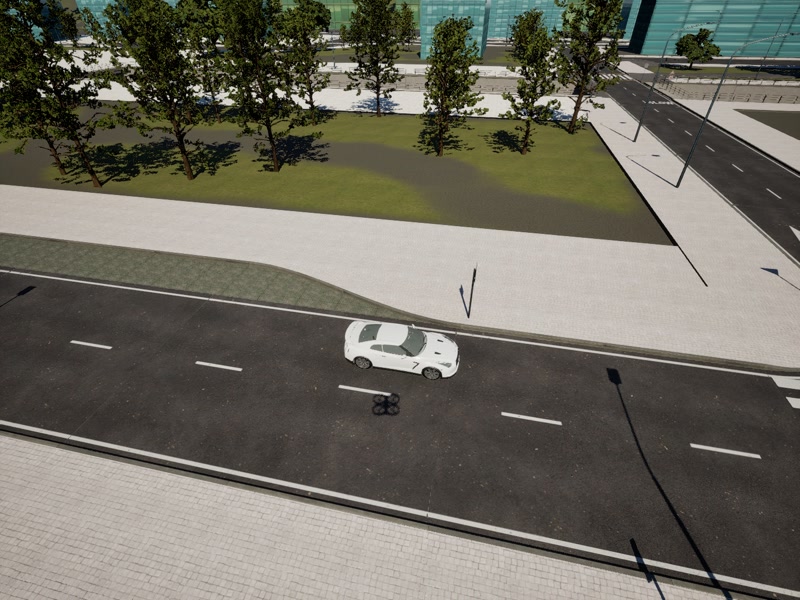}\\{\scriptsize $I_{t}$: $3.030$ s}\end{minipage}
\captionof{figure}{\textbf{Three-frame visual history.} Original onboard C1 frames, shown from oldest to newest. The labels give recorded acquisition times and their offsets from the current frame $I_t$. The policy interface order remains current, recent, older.}
\label{fig:temporal_scales}
\end{suppfigure}

\begin{suppfigure}
\includegraphics[width=0.90\linewidth]{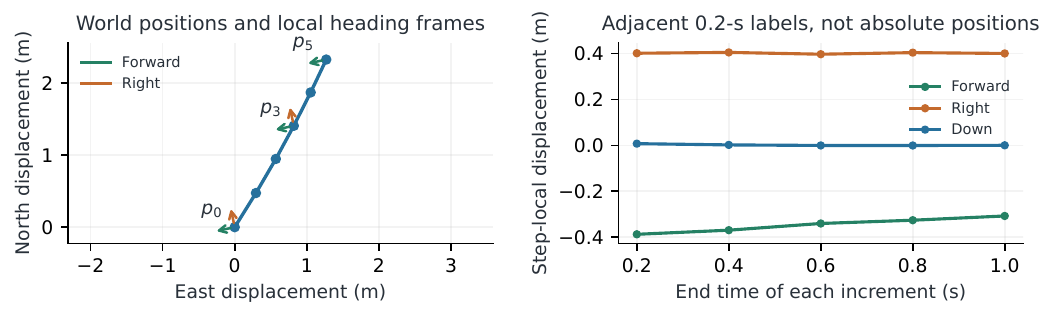}
\captionof{figure}{\textbf{Recorded motion and step-local incremental labels.} Six pose boundaries from the same standard C1 episode are aligned to a $0.2$-s grid by bounded interpolation of recorded control-tick poses (maximum bracketing gap below $0.15$ s). Left: horizontal world displacement with the heading-aligned forward/right axes at three boundaries. Right: five adjacent displacement labels. Each world difference is rotated by its own start yaw; inverse accumulation updates yaw after every step. Down is positive in NED. This illustrates the label convention, not controller tracking accuracy or a model prediction.}
\label{fig:coordinate_labels}
\end{suppfigure}

\subsection{Action Coverage and Stable-Tail Continuation}
\label{app:data_tail}

Finite-motion input samples capture available execution and endpoint adjustments up to the first qualifying stable H3 history. Subsequent holding frames provide future-label context rather than new inputs. Continuous-following samples demand complete recorded future windows. Temporal admissibility applies per sample, allowing extraction of partial episodes.

We use fully recorded futures whenever available. If only the terminal hold is missing, continuation requires evidence that the action and any necessary recovery have completed. The ordinary rule requires at least eight seconds of post-completion recording and a continuous two-second stable segment from the same episode. It allows up to $2.4$ seconds (12 future steps) per sample. We check stability from the donor segment through the end of the recording and at both splice boundaries. Only the future holding increments are continued. The RGB history and current state are unchanged, and the increments are not forced to zero.

Route-shortage exceptions are assessed per episode under frozen criteria. Continuation supplies supervision labels only and is not used in closed-loop tests.

Figure~\ref{fig:stable_tail} illustrates this stable-tail continuation rule.

\begin{figure}[!tp]
    \centering
    \includegraphics[width=0.98\linewidth]{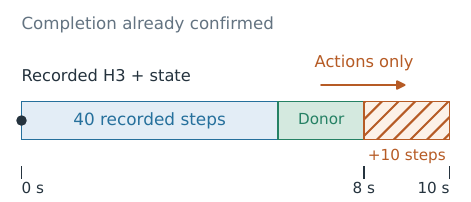}
    \caption{\textbf{Stable-tail continuation.} A ten-second label window contains eight seconds of recorded future motion and a two-second missing suffix. After endpoint and stability checks, the missing ten increments are sourced from a verified stable segment within the same episode. The original RGB history and state remain unchanged; we do not synthesize future images.}\label{fig:stable_tail}
\end{figure}

\subsection{Disturbed Trajectories and Language}
\label{app:data_language}

Strong-disturbance episodes are used only for training. Eligible windows may cover nominal motion, the disturbance, and subsequent recovery; labels follow the recorded motion.

Training and evaluation use the same canonical English instructions, specifying the shot and subject category without actor IDs, map names, target truth, or controller progress. Subject-relative directions are distinct from image-space thirds.

\subsection{Expert Reference Generation and Feedback}
\label{app:expert_control}

The collection expert tracks a subject-relative reference $\mathbf q(s)$, with shared progress $s\in[0,1]$ for coupled spatial changes. Orbit varies azimuth and, when requested, radius or height. Elevation varies height and may also change distance. Fly-over follows a longitudinal and lateral passage, while continuous following maintains the specified relative view. Spatial-endpoint shots slow progress when tracking error grows; fixed-duration shots follow a configured time profile.

Given target position $\mathbf p_{\mathrm s}$ and heading frame $Q$, the reference position is:
\begin{equation}
 \mathbf p_{\rm ref}(t)=\mathbf p_{\mathrm s}(t)+Q(t)\mathbf q(s(t)).
\label{eq:expert_reference}
\end{equation}
Feedforward combines subject translation with the derivative of relative motion and includes turning when configured. Position feedback, velocity damping, smoothing, and acceleration and speed limits produce the AirSim velocity command. Medium variations modify the reference smoothly before returning to baseline. Strong variants inject an external velocity disturbance, after which the same controller attempts recovery. Translational and UAV-yaw labels use recorded poses, not references or velocity commands.

Camera aim is computed from the line of sight to the subject. In ordinary modes, body yaw tracks horizontally and relative gimbal pitch adjusts elevation and framing; relative gimbal yaw remains fixed. F7 locks body yaw and adjusts relative gimbal yaw for horizontal thirds. The applied settings are smoothed and rate-limited before logging. The expert uses target truth for collection; the evaluated policy does not receive it.
Figure~\ref{fig:expert_reference_execution} connects the expert reference, recorded flight path, and measured supervision. Figure~\ref{fig:data_variants} highlights four variants of that pairing.

\begin{figure*}[!tp]
    \centering
    \includegraphics[width=0.98\linewidth]{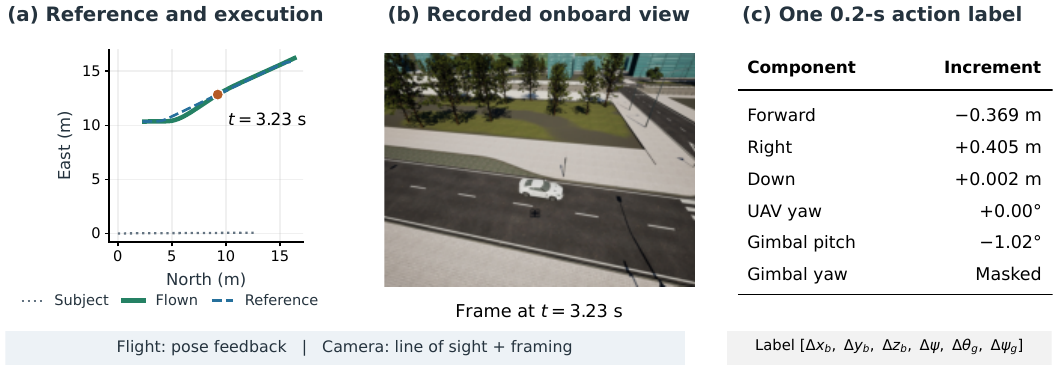}
    \caption{\textbf{From expert reference to measured supervision.} (a) The logged reference, flown path, and subject motion during a C1 side-follow and pull-away episode, shown in the world horizontal plane. (b) The onboard image at the marked time. Feedback keeps the UAV near the reference, while line-of-sight geometry aims the camera. (c) The first $0.2$-s increment computed by the training exporter at this frame, based on measured poses and applied gimbal settings. Gimbal yaw is masked for C1; angles are in degrees for readability.}\label{fig:expert_reference_execution}
\end{figure*}

\begin{suppfigure}
\begin{minipage}[t]{0.486\linewidth}\centering{\small\textbf{Standard clean}}\par\smallskip
\begin{minipage}[t]{0.326\linewidth}\centering\includegraphics[width=\linewidth]{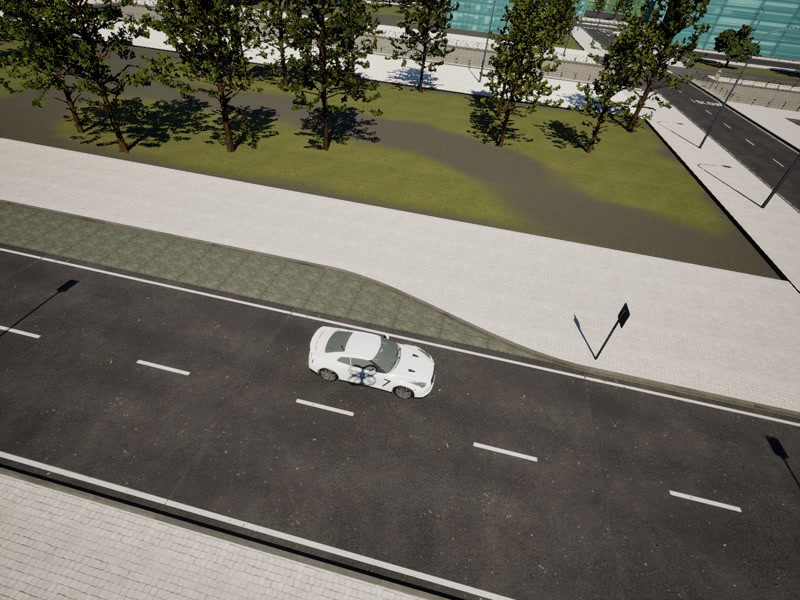}\\[-1pt]{\scriptsize Start\\$0.00$ s}\end{minipage}\hfill
\begin{minipage}[t]{0.326\linewidth}\centering\includegraphics[width=\linewidth]{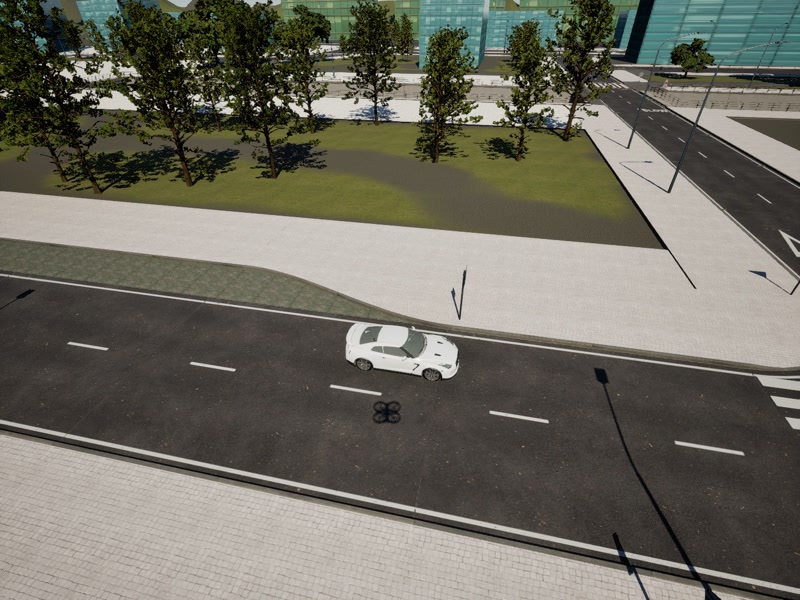}\\[-1pt]{\scriptsize Executing\\$3.20$ s}\end{minipage}\hfill
\begin{minipage}[t]{0.326\linewidth}\centering\includegraphics[width=\linewidth]{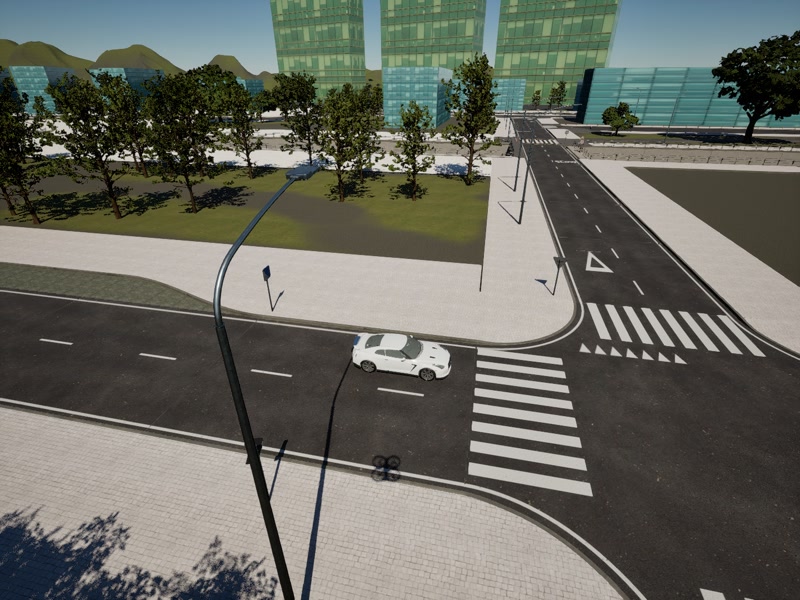}\\[-1pt]{\scriptsize Post-completion\\$8.40$ s}\end{minipage}\end{minipage}\hfill
\begin{minipage}[t]{0.486\linewidth}\centering{\small\textbf{Single-axis: height}}\par\smallskip
\begin{minipage}[t]{0.326\linewidth}\centering\includegraphics[width=\linewidth]{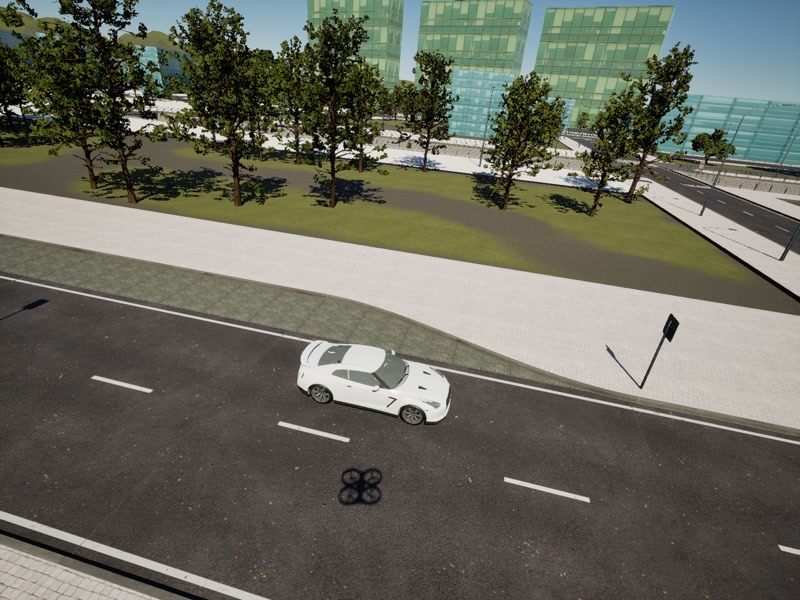}\\[-1pt]{\scriptsize Start\\$0.00$ s}\end{minipage}\hfill
\begin{minipage}[t]{0.326\linewidth}\centering\includegraphics[width=\linewidth]{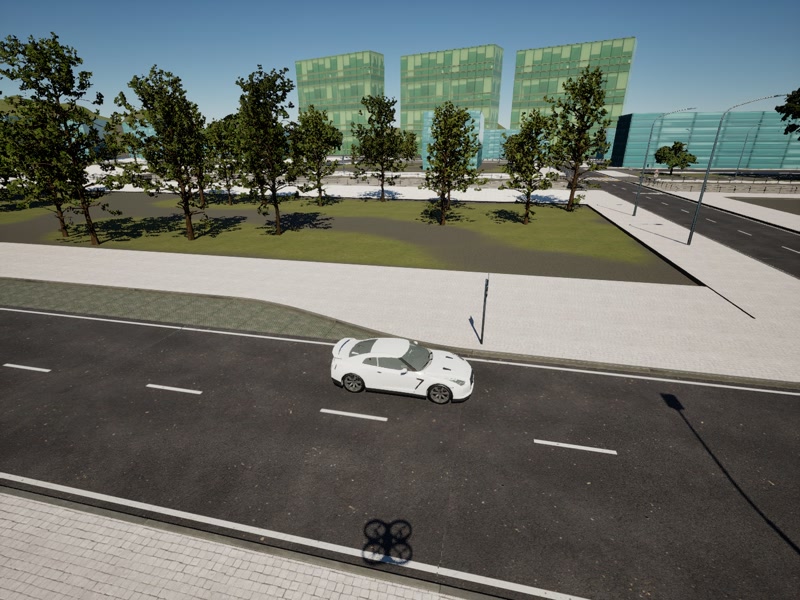}\\[-1pt]{\scriptsize Executing\\$3.00$ s}\end{minipage}\hfill
\begin{minipage}[t]{0.326\linewidth}\centering\includegraphics[width=\linewidth]{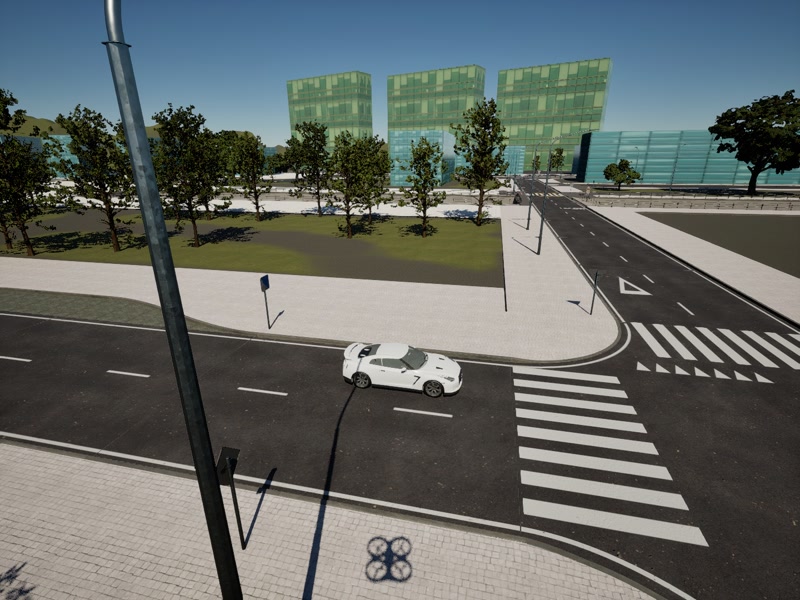}\\[-1pt]{\scriptsize Post-completion\\$7.99$ s}\end{minipage}\end{minipage}
\par\medskip
\begin{minipage}[t]{0.486\linewidth}\centering{\small\textbf{Dual-axis: action rate + lateral}}\par\smallskip
\begin{minipage}[t]{0.326\linewidth}\centering\includegraphics[width=\linewidth]{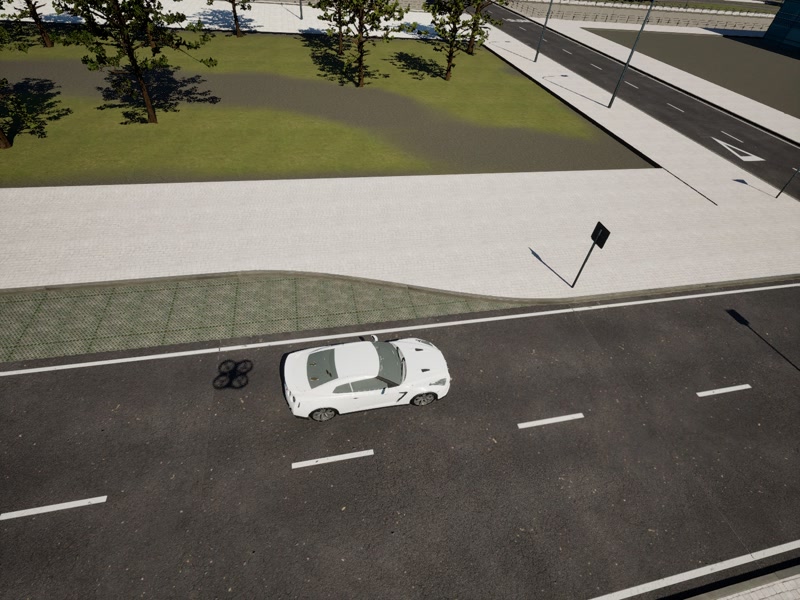}\\[-1pt]{\scriptsize Start\\$0.00$ s}\end{minipage}\hfill
\begin{minipage}[t]{0.326\linewidth}\centering\includegraphics[width=\linewidth]{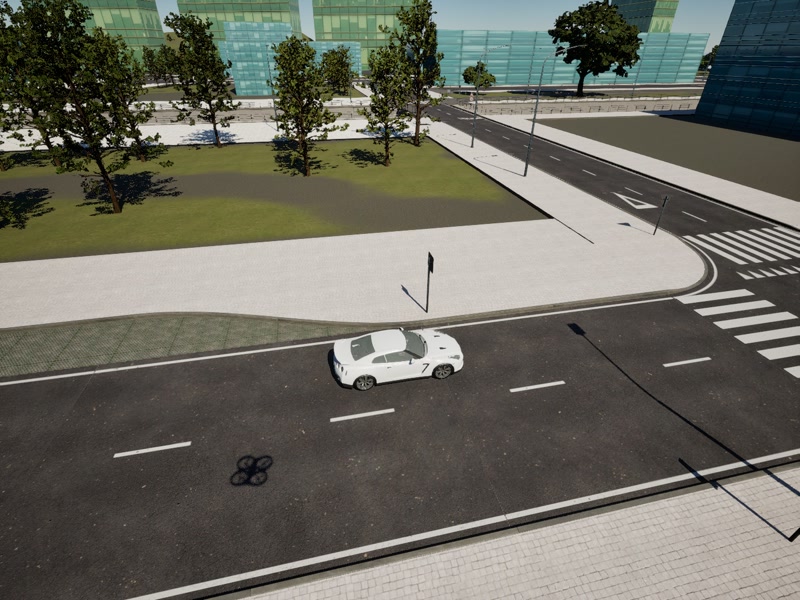}\\[-1pt]{\scriptsize Executing\\$3.22$ s}\end{minipage}\hfill
\begin{minipage}[t]{0.326\linewidth}\centering\includegraphics[width=\linewidth]{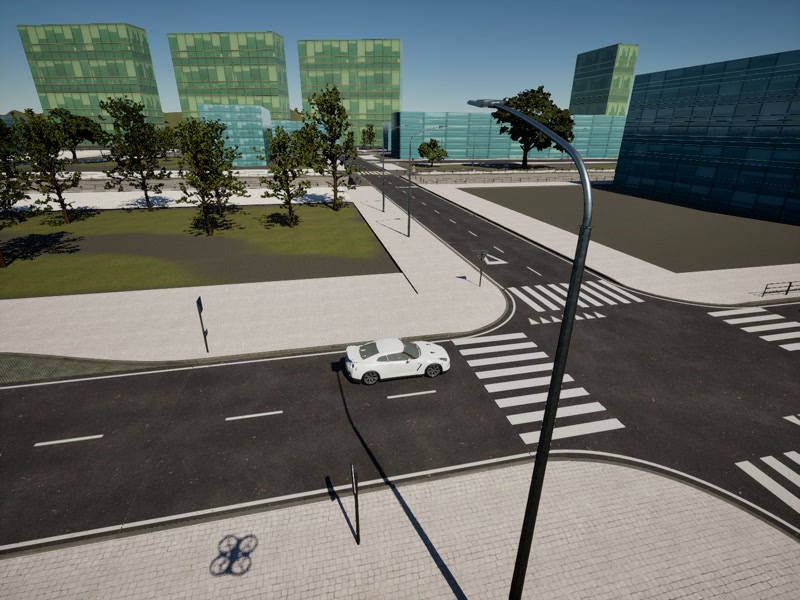}\\[-1pt]{\scriptsize Post-completion\\$8.42$ s}\end{minipage}\end{minipage}\hfill
\begin{minipage}[t]{0.486\linewidth}\centering{\small\textbf{Strong lateral injection}}\par\smallskip
\begin{minipage}[t]{0.326\linewidth}\centering\includegraphics[width=\linewidth]{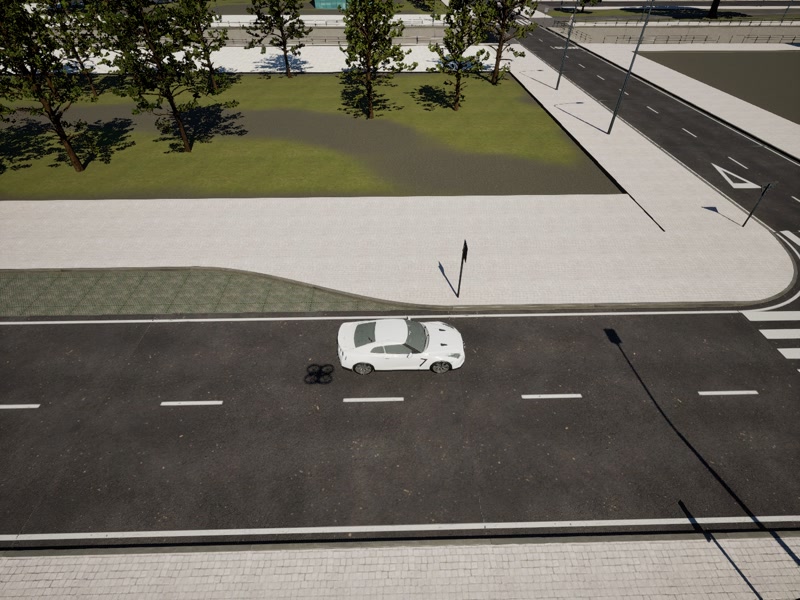}\\[-1pt]{\scriptsize Injection\\$2.93$ s}\end{minipage}\hfill
\begin{minipage}[t]{0.326\linewidth}\centering\includegraphics[width=\linewidth]{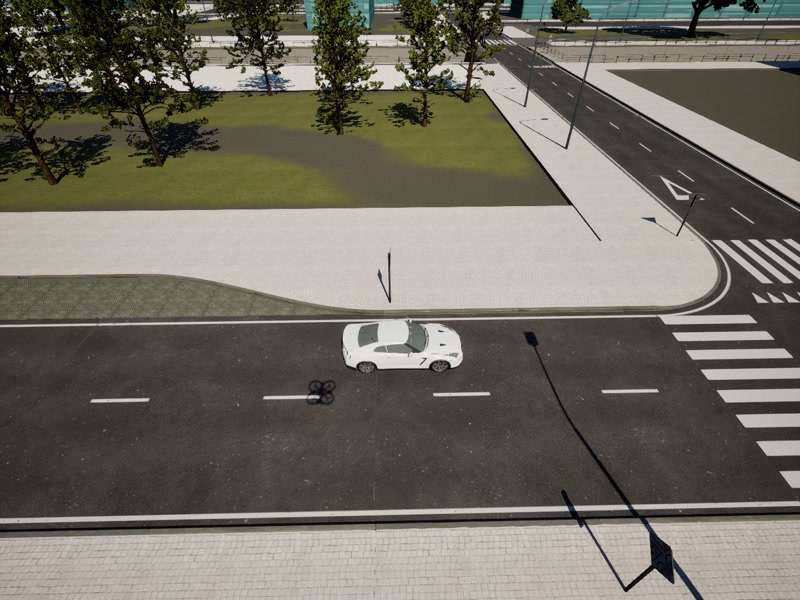}\\[-1pt]{\scriptsize Post-release\\$4.45$ s}\end{minipage}\hfill
\begin{minipage}[t]{0.326\linewidth}\centering\includegraphics[width=\linewidth]{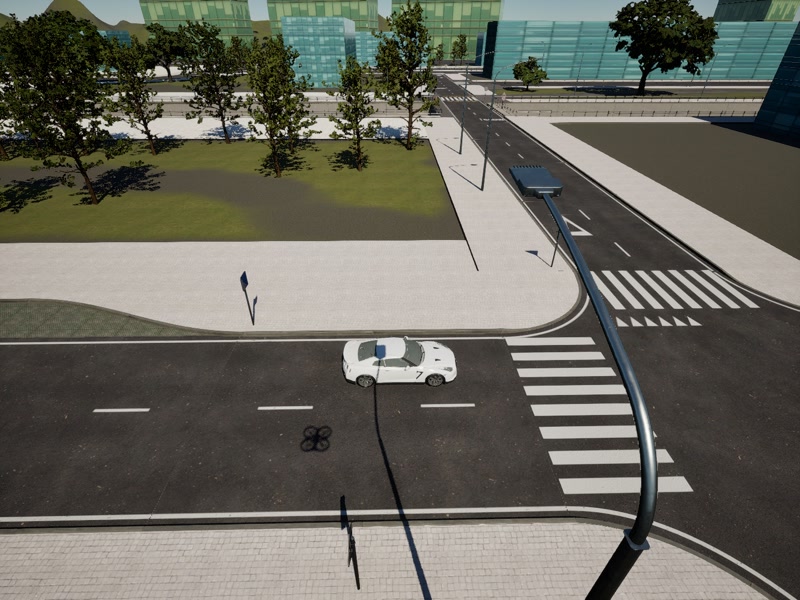}\\[-1pt]{\scriptsize Post-completion\\$7.63$ s}\end{minipage}\end{minipage}
\captionof{figure}{\textbf{Four recorded variants of C1 (side-follow and pull-away).} The standard clean demonstration and two medium-variation examples show the start, execution, and post-completion view. The strong-recovery example shows the velocity injection, post-release flight, and post-completion view. Times come from acquisition metadata. Baseline distance, height, and speed also vary, so the examples are not a single-variable comparison.}
\label{fig:data_variants}
\end{suppfigure}

\endgroup

% !TeX root = ../../main.tex
% Interfaces and training settings are tied to the actually evaluated checkpoints.
\begingroup

\section{Model Interfaces and Task Adaptation}
\label{app:model_adaptation}

\subsection{Common Task Contract}
\label{app:model_contract}

All policies receive canonical English instructions, genuine H3 RGB history, and causal UAV-only proprioception. Each model retains its own image processor and task-specific normalization, maintaining consistency between training and inference. The action interface represents heading-aligned translations and angular increments (UAV and gimbal) in metres and radians at $0.2$-second intervals. Models do not receive the target position, target velocity, relative geometric truth, or expert action progress.

The common supervision export contains 50 future action steps per sample. Prediction horizons vary by model (Table~\ref{tab:model_prediction_horizons}); closed-loop tests execute only the first ten predicted steps per cycle, subject to early stopping.

\begin{table*}[!tp]
    \centering
    \caption{Task-adapted policy interfaces. All models use a ten-step execution prefix, irrespective of prediction length. DreamZero denotes the task-specific 5B adaptation.}
    \label{tab:model_prediction_horizons}
    \small
    \begin{tabular}{@{}lp{0.65\linewidth}r@{}}
        \toprule
        Policy & Initialization and task adaptation & \shortstack[r]{Prediction horizon\\(steps)} \\
        \midrule
        PI0 & Pretrained flow-based VLA with the photography state/action interface & 50 \\
        PI0.5 & Pretrained VLA with the photography interface and selected state/action statistics & 50 \\
        SmolVLA & Pretrained VLA with history and masked incremental actions & 50 \\
        WALL-OSS 0.5 & Original base policy, native action head, and a 32-step task view & 32 \\
        OpenVLA-OFT & OpenVLA-7B with an OFT continuous L1 action head and proprioceptive projection & 10 \\
        ACT & Language-conditioned Transformer/CVAE chunk policy with pretrained visual features & 50 \\
        Diffusion Policy & Language-conditioned temporal U-Net action diffusion with pretrained visual features & 50 \\
        Cosmos Policy & Original Cosmos-Policy-ALOHA-Predict2-2B policy adapted to photography & 50 \\
        DreamZero & Wan2.2-TI2V-5B backbone with rank-16 LoRA and new task layers & 24 \\
        \bottomrule
    \end{tabular}
\end{table*}

\subsection{Adaptation-Specific Distinctions}
\label{app:model_distinctions}

OpenVLA-OFT uses a continuous L1 regression head, in contrast to the discrete autoregressive interface of the original OpenVLA. We adapt the action head and proprioceptive projection to the aerial interface. WALL retains its native head and uses the first 32 steps of the shared labels. The DreamZero-derived model freezes the 5B backbone and trains rank-16 LoRA modules and new image, state, and action layers. It uses the first 24 label steps and, where available, aligned real future-video supervision. Cosmos Policy starts from a pretrained robot-policy checkpoint.

ACT and Diffusion Policy are language-conditioned action policies without a pretrained VLA backbone. Both use ImageNet-pretrained ResNet18 visual features and a frozen MiniLM language encoder. ACT predicts action chunks using a Transformer/CVAE, whereas Diffusion Policy jointly denoises temporal action sequences. Both learn the cinematography policy on the shared training split.

When used as auxiliary targets by video-derived policies, future images are not inference inputs. Normalization statistics use the training split only. Native losses across different prediction horizons and normalizations are not directly comparable.

\subsection{Training Settings and Evaluated Checkpoints}
\label{app:model_training}

Table~\ref{tab:model_training_settings} summarizes the training configurations and evaluated checkpoints. Training samples are shuffled each epoch. Offline validation uses a fixed subset of 3,280 samples with fully recorded future actions (eight per map--semantic group). Checkpoint selection does not use continued tails or closed-loop test results.

\begin{table*}[!tp]
\centering
\small
\caption{Training configurations and evaluated checkpoints. Batch size is global, checkpoint steps count optimizer updates, and learning-rate arrows denote continuations at a lower learning rate.}
\label{tab:model_training_settings}
\begin{tabular}{@{}llrrl@{}}
\toprule
Policy & Updated components & Batch & Evaluated step & Peak / continued LR \\
\midrule
PI0 & Full policy & 64 & 15,000 & $10^{-5}$ \\
PI0.5 & Full policy, state-Q/action-STD & 48 & 23,000 & $10^{-5}$ \\
SmolVLA & Full policy & 32 & 48,000 & $10^{-5}\rightarrow10^{-6}$ \\
WALL-OSS 0.5 & Full policy & 32 & 28,000 & $10^{-5}$ \\
OpenVLA-OFT & Full VLM, new head and state projection & 64 & 18,866 & $10^{-5}$ \\
ACT & Action policy and visual encoder & 256 & 24,500 & $10^{-4}\rightarrow10^{-5}$ \\
Diffusion Policy & Action policy and visual encoder & 256 & 18,500 & $10^{-4}$ \\
Cosmos Policy & Policy network; text/VAE frozen & 64 & 18,000 & $10^{-5}$ \\
DreamZero & LoRA and new task layers & 48 & 25,156 & $10^{-4}$ \\
\bottomrule
\end{tabular}
\end{table*}

\noindent\textbf{Optimization.}
All models train with AdamW, learning-rate warmup, and cosine decay. VLA policies warm up over 1,000 steps; other policies warm up over 500 steps. ACT applies a lower learning rate to its visual encoder than to its action policy. The SmolVLA and ACT continuation phases use the reduced rates listed in Table~\ref{tab:model_training_settings}.

\noindent\textbf{Preprocessing.}
The common camera stream is $1440\times1080$ RGB with a $90^\circ$ horizontal field of view. Each model applies its own image processor without reordering the history. ACT, Diffusion Policy, and PI0.5 use $224\times224$ inputs; ACT and Diffusion Policy use ImageNet normalization. PI0, SmolVLA, and ACT normalize state and action using mean and standard deviation. PI0.5 uses state quantiles and action mean--standard-deviation statistics. WALL uses clipped quantiles, and Diffusion Policy applies min--max scaling. Action normalization matches each model's trained prediction horizon.
\endgroup

% !TeX root = ../../main.tex
% New catalogue and case construction are extracted from the frozen evaluation records.
\begingroup

\section{Semantic Catalogue and Fixed Test Construction}
\label{app:semantic_catalog}

\subsection{Concrete Semantics and Analysis Groups}
\label{app:semantic_groups}

Table~\ref{tab:semantic_catalog} lists all 41 concrete cinematography semantics: ten \emph{Follow}, fifteen \emph{Orbit}, six \emph{Elevation}, four \emph{Fly-over}, and six compositional semantics. The catalogue specifies the subject state and execution regime for each semantic. It comprises 26 static-subject and 15 moving-subject semantics, partitioned into eight continuous-following and 33 finite-motion semantics.

We fixed the single-/joint-requirement grouping prior to comparing models. A single requirement specifies one primary motion or relative viewpoint. A joint requirement couples explicit spatial changes, subject following with a viewpoint change, or a motion with thirds framing. For example, F5 and F6 couple subject following with distance change, F7 couples following with image-space framing, and E5 couples rising with vertical thirds framing. The frozen grouping contains 20 single- and 21 joint-requirement semantics.

In the English instructions, SUBJECT acts as a placeholder for \emph{vehicle}, \emph{truck}, or \emph{person}. Directions are subject-relative, orbit direction is defined from above, and thirds refer to image-space regions. Semantics O3--O6 and C4/C5 do not constrain the orbit direction.

Figure~\ref{fig:shot_family_gallery} illustrates the five shot families using expert demonstration frames with overlaid schematic motion trajectories.

\subsection{Construction of Fixed Test Cases}
\label{app:test_construction}

Each semantic is instantiated on two suitable maps. Within each map, two training-seen semantic--subject pairings and one unseen pairing each contribute ten test cases with case-specific initial conditions and scene settings. This yields $41\times2\times3\times10=2,460$ cases: 1,640 seen-pair and 820 unseen-pair cases. Unseen refers to the pairing, not to a new semantic or map. All models use the same cases and budgets.

Case allocation is balanced across semantics rather than maps. Subject and map selection ensure feasible route lengths and viewpoints. All task-attribute analyses share this single fixed test set.

\subsection{Translational Progress References}
\label{app:progress_references}

For a required radius or height axis $a$, let $x_a(t)$ be its measured value and $\sigma_a$ be $+1$ for an increase or $-1$ for a decrease. Table~\ref{tab:linear_progress_references} gives the milestone amplitude $A_a$ and terminal bounds $[L_a,U_a]$ for the signed change $\sigma_a[x_a(t)-x_a(0)]$. These thresholds are used only for closed-loop scoring (Appendix~\ref{app:metric_parameters}).

\noindent\begin{minipage}{\linewidth}
\centering
\small
\captionsetup{hypcap=false}
\captionof{table}{Reference amplitudes and terminal bounds for radius and height changes (metres).}
\label{tab:linear_progress_references}
\begin{tabular}{@{}llcrr@{}}
\toprule
Semantic & Axis & $\sigma_a$ & $A_a=L_a$ & $U_a$ \\
\midrule
C1 & Radius & $+$ & 2.9135 & 6.8395 \\
C2 & Height & $+$ & 3.2730 & 13.8708 \\
C3 & Radius & $+$ & 2.9357 & 19.2243 \\
C3 & Height & $+$ & 3.1669 & 14.1711 \\
C4 & Radius & $+$ & 1.0922 & 4.7635 \\
C4 & Height & $+$ & 2.2620 & 12.0988 \\
C5 & Radius & $-$ & 2.9031 & 12.0000 \\
C5 & Height & $-$ & 4.7628 & 10.8448 \\
E1 & Height & $+$ & 1.9829 & 13.7967 \\
E2 & Height & $-$ & 2.0890 & 11.2257 \\
E3 & Radius & $+$ & 3.0109 & 14.0788 \\
E3 & Height & $+$ & 2.6803 & 13.8947 \\
E4 & Radius & $-$ & 5.6394 & 14.5946 \\
E4 & Height & $-$ & 1.9237 & 11.0063 \\
E5/lower\_third & Height & $+$ & 2.1876 & 10.7116 \\
E5/upper\_third & Height & $+$ & 1.6253 & 5.3364 \\
F5 & Radius & $-$ & 5.5630 & 12.6329 \\
F6 & Radius & $+$ & 2.8804 & 18.9858 \\
O3 & Radius & $-$ & 4.6106 & 11.8288 \\
O4 & Radius & $+$ & 1.7177 & 17.5359 \\
O5 & Height & $+$ & 2.0785 & 6.3432 \\
O6 & Height & $-$ & 1.4668 & 9.1708 \\
\bottomrule
\end{tabular}
\end{minipage}\par
\endgroup

% The catalogue spans the full page width and breaks between rows.
\onecolumn
% Generated from the frozen instruction catalog and semantic grouping manifest.
\begingroup
\small
\setlength{\tabcolsep}{4pt}
\setlength{\LTcapwidth}{\textwidth}
\setlength{\LTpre}{0pt}
\begin{longtable}{@{}p{0.24\textwidth}p{0.18\textwidth}p{0.54\textwidth}@{}}
\caption{All 41 concrete cinematography semantics and their frozen analysis groups. SUBJECT is replaced by a subject category, not an actor identity.}\label{tab:semantic_catalog}\\
\toprule
Semantic & Subject / execution / requirements & Canonical instruction \\
\midrule
\endfirsthead
\caption[]{Concrete cinematography semantics and frozen analysis groups (continued).}\\
\toprule
Semantic & Subject / execution / requirements & Canonical instruction \\
\midrule
\endhead
\bottomrule
\endfoot
F1/base & moving / continuous / single & Follow the moving SUBJECT from behind and maintain a steady distance. \\
F2/base & moving / continuous / single & Stay ahead of the moving SUBJECT and film its front while maintaining a steady distance. \\
F3/left & moving / continuous / single & Follow alongside the moving SUBJECT on its left and maintain a steady lateral distance. \\
F3/right & moving / continuous / single & Follow alongside the moving SUBJECT on its right and maintain a steady lateral distance. \\
F4/rear\_\allowbreak{}left & moving / continuous / single & Follow the moving SUBJECT from its rear-left side. \\
F4/rear\_\allowbreak{}right & moving / continuous / single & Follow the moving SUBJECT from its rear-right side. \\
F5/base & moving / finite / joint & Follow the moving SUBJECT while smoothly moving closer, then maintain the final distance. \\
F6/base & moving / finite / joint & Follow the moving SUBJECT while smoothly pulling away, then maintain the final distance. \\
F7/left\_\allowbreak{}third & moving / continuous / joint & Follow the moving SUBJECT, move it to the left third of the frame, and keep it there. \\
F7/right\_\allowbreak{}third & moving / continuous / joint & Follow the moving SUBJECT, move it to the right third of the frame, and keep it there. \\
O1/base & static / finite / single & Orbit the stationary SUBJECT clockwise for one full circle, then hold the final view. \\
O2/base & static / finite / single & Orbit the stationary SUBJECT counterclockwise for one full circle, then hold the final view. \\
O3/base & static / finite / joint & Orbit the stationary SUBJECT once while moving closer, then hold the final view. \\
O4/base & static / finite / joint & Orbit the stationary SUBJECT once while pulling away, then hold the final view. \\
O5/base & static / finite / joint & Orbit the stationary SUBJECT once while rising, then hold the final view. \\
O6/base & static / finite / joint & Orbit the stationary SUBJECT once while descending, then hold the final view. \\
O7/base & moving / finite / joint & Orbit the moving SUBJECT clockwise for one full circle while tracking it, then hold the final relative view. \\
O8/front\_\allowbreak{}to\_\allowbreak{}left\_\allowbreak{}ccw\_\allowbreak{}q & static / finite / single & Make a counterclockwise quarter orbit around the stationary SUBJECT from its front to its left side, then hold. \\
O8/front\_\allowbreak{}to\_\allowbreak{}right\_\allowbreak{}cw\_\allowbreak{}q & static / finite / single & Make a clockwise quarter orbit around the stationary SUBJECT from its front to its right side, then hold. \\
O8/left\_\allowbreak{}to\_\allowbreak{}right\_\allowbreak{}ccw\_\allowbreak{}h & static / finite / single & Make a counterclockwise half orbit around the stationary SUBJECT from its left side to its right side, then hold. \\
O8/left\_\allowbreak{}to\_\allowbreak{}right\_\allowbreak{}cw\_\allowbreak{}h & static / finite / single & Make a clockwise half orbit around the stationary SUBJECT from its left side to its right side, then hold. \\
O8/rear\_\allowbreak{}to\_\allowbreak{}front\_\allowbreak{}ccw\_\allowbreak{}h & static / finite / single & Make a counterclockwise half orbit around the stationary SUBJECT from its rear to its front, then hold. \\
O8/rear\_\allowbreak{}to\_\allowbreak{}front\_\allowbreak{}cw\_\allowbreak{}h & static / finite / single & Make a clockwise half orbit around the stationary SUBJECT from its rear to its front, then hold. \\
O8/right\_\allowbreak{}to\_\allowbreak{}left\_\allowbreak{}ccw\_\allowbreak{}h & static / finite / single & Make a counterclockwise half orbit around the stationary SUBJECT from its right side to its left side, then hold. \\
O8/right\_\allowbreak{}to\_\allowbreak{}left\_\allowbreak{}cw\_\allowbreak{}h & static / finite / single & Make a clockwise half orbit around the stationary SUBJECT from its right side to its left side, then hold. \\
E1/base & static / finite / single & Rise while filming the stationary SUBJECT, keep the horizontal position steady, then hold. \\
E2/base & static / finite / single & Descend while filming the stationary SUBJECT, keep the horizontal position steady, then hold. \\
E3/base & static / finite / joint & Rise and pull away from the stationary SUBJECT, then hold the final view. \\
E4/base & static / finite / joint & Descend and move closer to the stationary SUBJECT, then hold the final view. \\
E5/upper\_\allowbreak{}third & static / finite / joint & Rise while filming the stationary SUBJECT, move it to the upper third of the frame, then hold. \\
E5/lower\_\allowbreak{}third & static / finite / joint & Rise while filming the stationary SUBJECT, move it to the lower third of the frame, then hold. \\
V1/base & static / finite / single & Fly over the stationary SUBJECT from its rear to its front, keep it visible, then hold the final view. \\
V2/base & static / finite / single & Fly over the stationary SUBJECT from its left side to its right side, then hold the final view. \\
V3/rear\_\allowbreak{}left\_\allowbreak{}to\_\allowbreak{}front\_\allowbreak{}right & static / finite / joint & Fly diagonally over the stationary SUBJECT from its rear-left to its front-right, keep it visible, then hold. \\
V3/rear\_\allowbreak{}right\_\allowbreak{}to\_\allowbreak{}front\_\allowbreak{}left & static / finite / joint & Fly diagonally over the stationary SUBJECT from its rear-right to its front-left, keep it visible, then hold. \\
C1/base & moving / finite / joint & Follow alongside the moving SUBJECT while pulling away, then maintain the final relative view. \\
C2/base & moving / finite / joint & Follow the moving SUBJECT from a rear-oblique view while rising, then maintain the final relative view. \\
C3/base & moving / finite / joint & Follow the moving SUBJECT while rising and pulling away, then maintain the final relative view. \\
C4/base & static / finite / joint & Orbit the stationary SUBJECT once while rising and pulling away, then hold the final view. \\
C5/base & static / finite / joint & Orbit the stationary SUBJECT once while descending and moving closer, then hold the final view. \\
C6/base & moving / finite / joint & Fly over the moving SUBJECT from its rear to its front along its direction of travel, keep it visible, then track from ahead. \\
\end{longtable}
\endgroup

% !TeX root = ../../main.tex
% Keep each heading with its opening figure without forcing a page break.
\noindent\begin{minipage}{\textwidth}
\section{Qualitative Evidence}
\label{app:visual_examples}

\subsection{Expert Previews of the Shot Families}
\label{app:expert_visuals}

\noindent\makebox[\textwidth][c]{\begin{minipage}{0.95\textwidth}\centering
\makebox[0.193\linewidth][c]{\scriptsize\textbf{F1 / Follow}}\hfill
\makebox[0.193\linewidth][c]{\scriptsize\textbf{O7 / Orbit}}\hfill
\makebox[0.193\linewidth][c]{\scriptsize\textbf{E2 / Elevation}}\hfill
\makebox[0.193\linewidth][c]{\scriptsize\textbf{V1 / Fly-over}}\hfill
\makebox[0.193\linewidth][c]{\scriptsize\textbf{C1 / Composite}}\par
\vspace{3pt}
\begin{minipage}[c]{0.193\linewidth}\centering\includegraphics[width=\linewidth]{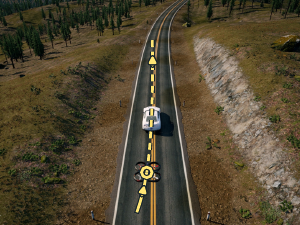}\end{minipage}\hfill
\begin{minipage}[c]{0.193\linewidth}\centering\includegraphics[width=\linewidth]{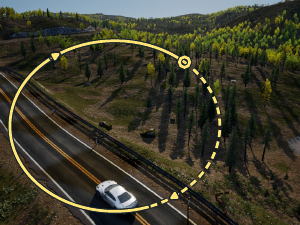}\end{minipage}\hfill
\begin{minipage}[c]{0.193\linewidth}\centering\includegraphics[width=\linewidth]{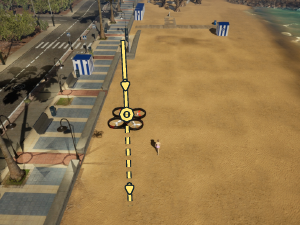}\end{minipage}\hfill
\begin{minipage}[c]{0.193\linewidth}\centering\includegraphics[width=\linewidth]{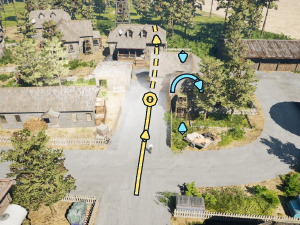}\end{minipage}\hfill
\begin{minipage}[c]{0.193\linewidth}\centering\includegraphics[width=\linewidth]{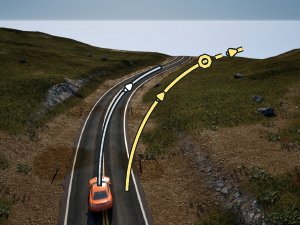}\end{minipage}\par
\vspace{2pt}
\begin{minipage}[c]{0.193\linewidth}\centering\includegraphics[width=\linewidth]{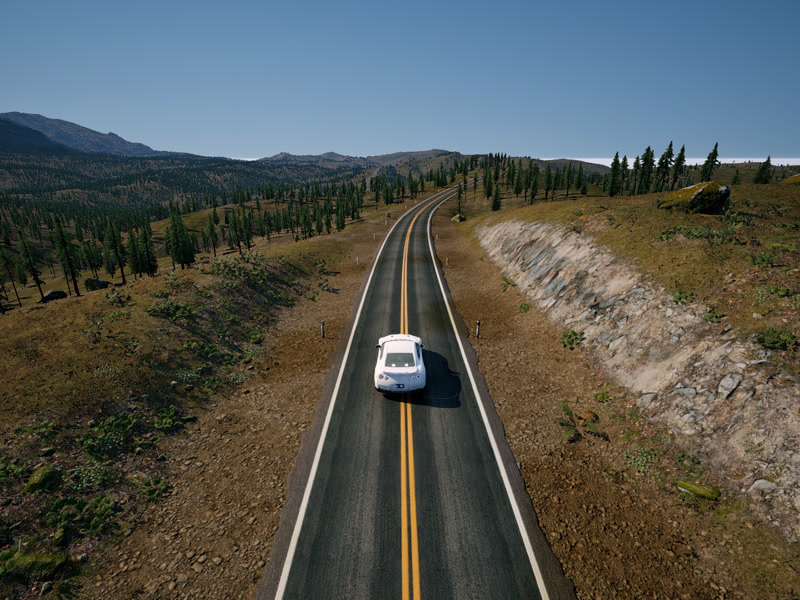}\end{minipage}\hfill
\begin{minipage}[c]{0.193\linewidth}\centering\includegraphics[width=\linewidth]{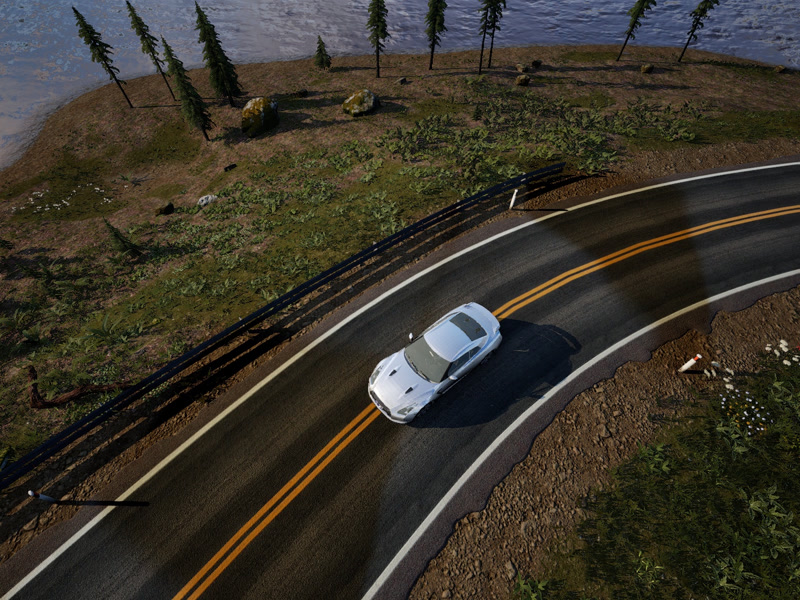}\end{minipage}\hfill
\begin{minipage}[c]{0.193\linewidth}\centering\includegraphics[width=\linewidth]{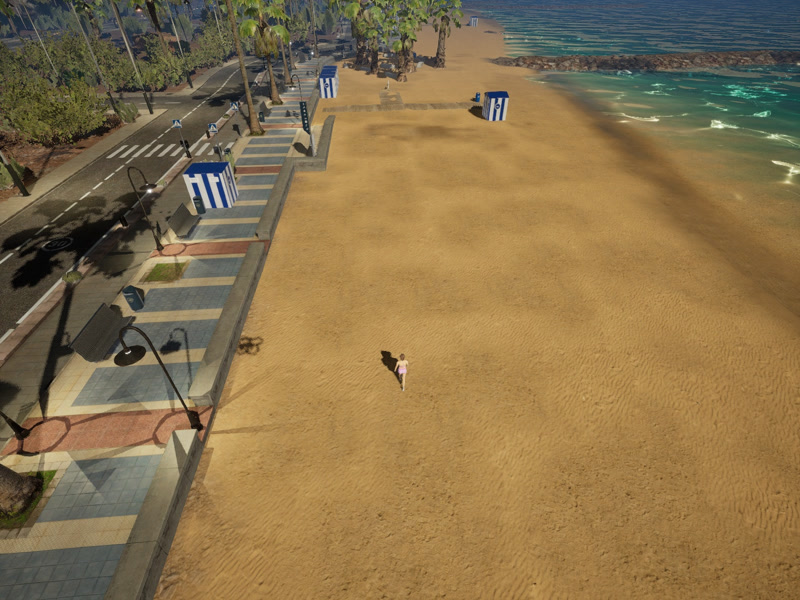}\end{minipage}\hfill
\begin{minipage}[c]{0.193\linewidth}\centering\includegraphics[width=\linewidth]{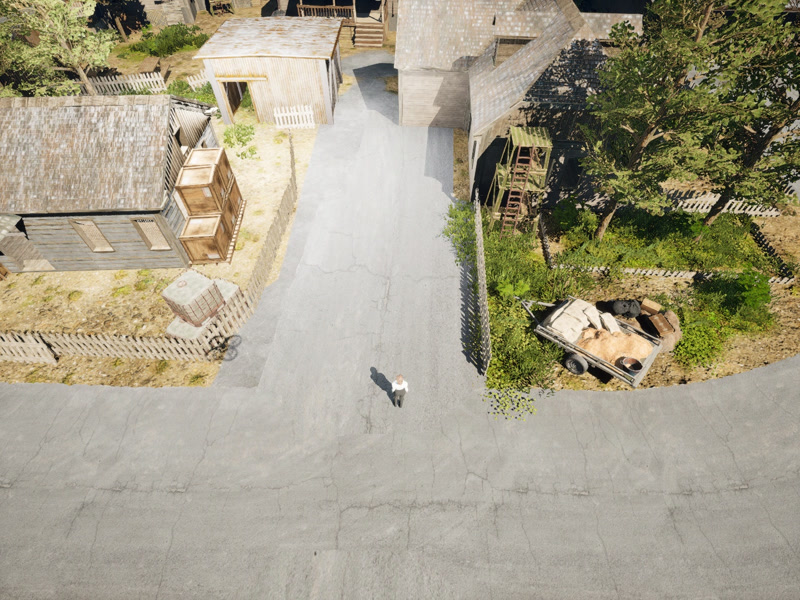}\end{minipage}\hfill
\begin{minipage}[c]{0.193\linewidth}\centering\includegraphics[width=\linewidth]{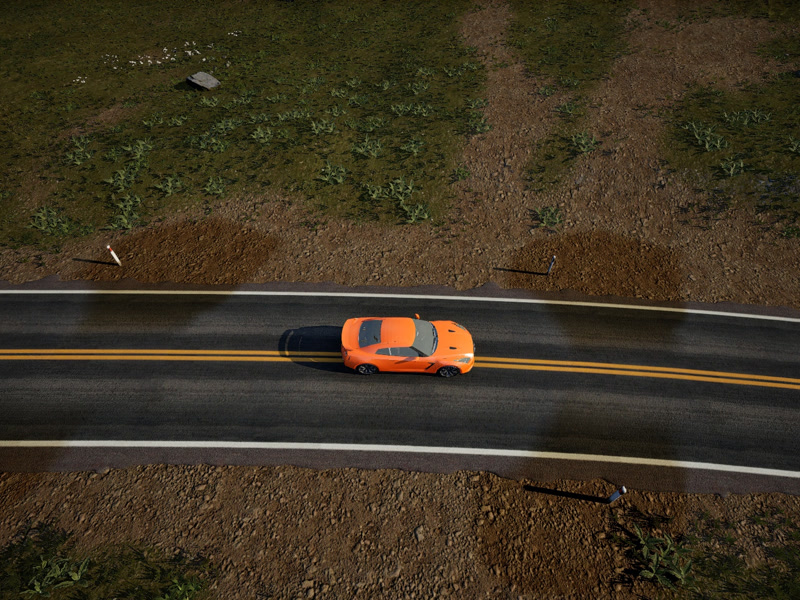}\end{minipage}\par
\captionof{figure}{\textbf{Five expert shot families.} Top: schematic motion overlaid on observer frames. The ring marks the UAV. Gold solid and dashed lines show its path before and after the shown instant. Blue arrows show V1 heading; the white arrow shows the C1 subject course. Bottom: onboard frames from the same demonstrations.}\label{fig:shot_family_gallery}
\end{minipage}}\par

\end{minipage}\par
\vspace{10pt}

\noindent\begin{minipage}{\textwidth}
\subsection{Matched Cases: What In-Frame Scores Miss}
\label{app:policy_visuals}

\noindent In each CityRoads case below, the compared policies share the initialization, instruction, and budget. Boxes identify the visible vehicle; traces and scores come from the recorded tests. Times below the images are nominal decision times. The cases illustrate execution errors rather than overall rankings.
\smallskip

\noindent\begin{minipage}{\textwidth}
\centering
\includegraphics[width=0.95\textwidth]{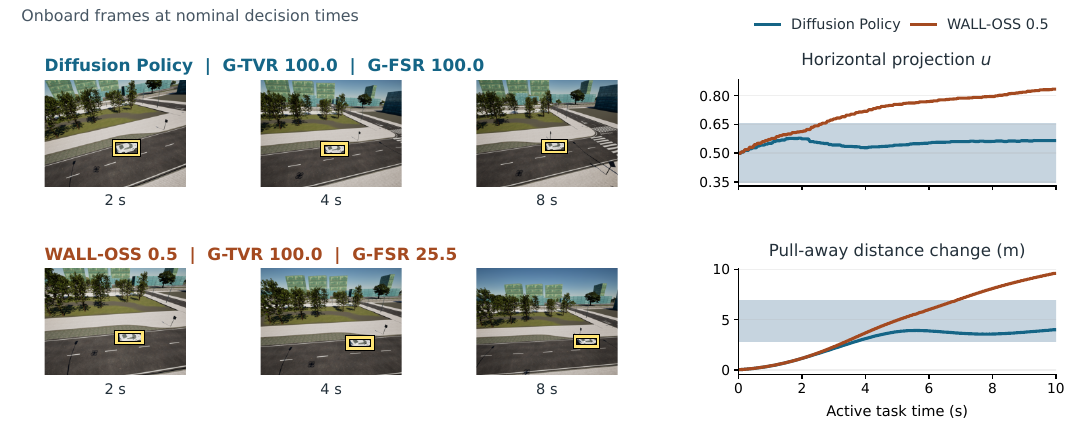}
\captionof{figure}{\textbf{C1: Target retention does not guarantee correct framing.} Both policies keep the subject within the geometric field of view. With WALL-OSS 0.5, the subject projection drifts rightward and the pull-away exceeds the accepted terminal distance change. Diffusion Policy stays near the requested horizontal framing position. The upper band marks horizontal framing tolerance; the lower band gives the terminal distance-change interval.}
\label{fig:focused_c1}
\end{minipage}
\end{minipage}\par
\vspace{8pt}

\noindent\begin{minipage}{\textwidth}
\centering
\includegraphics[width=0.89\textwidth]{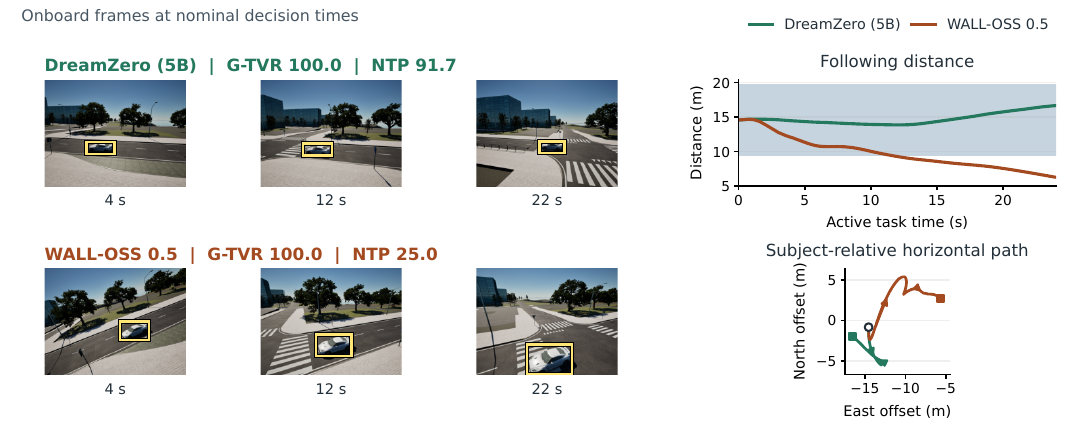}
\captionof{figure}{\textbf{F3: Target retention does not guarantee sustained following.} WALL-OSS 0.5 steadily closes in on the moving subject, whereas DreamZero maintains the intended distance. DreamZero passes the joint following criteria in 11 of 12 evaluation windows, compared to 3 of 12 for WALL-OSS 0.5. The lower panel displays the measured subject-relative paths.}
\label{fig:focused_f3}
\end{minipage}\par
\vspace{8pt}

\noindent\begin{minipage}{\textwidth}
\subsection{Static Subjects: Orbit and Rise}
\label{app:orbit_visuals}

\noindent\begin{minipage}{\textwidth}
\centering
\includegraphics[width=0.89\textwidth]{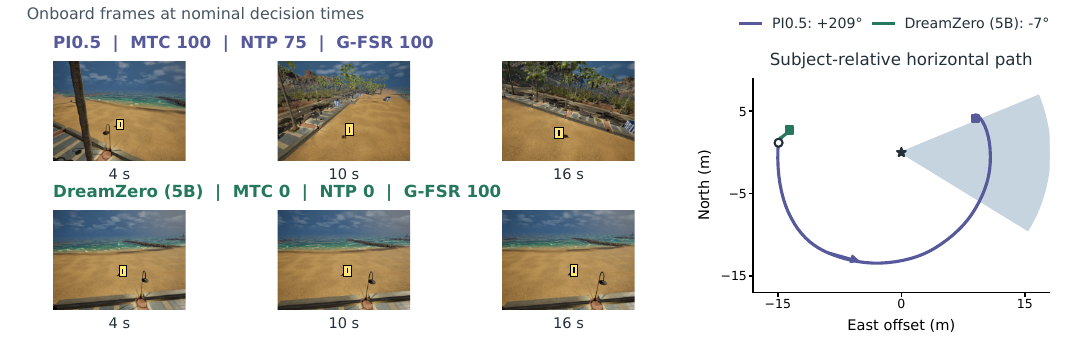}
\captionof{figure}{\textbf{O8 on Coast: Correct framing does not verify orbit completion.} On this unseen semantic--subject pairing, both policies successfully keep the stationary person in the framing region. PI0.5 orbits counterclockwise as requested but reaches approximately $209^\circ$, exceeding the accepted half-orbit interval $[153,207]^\circ$. DreamZero remains near the starting azimuth. Shading denotes the accepted terminal angular interval.}
\label{fig:focused_orbit}
\end{minipage}
\end{minipage}\par
\vspace{8pt}

\noindent\begin{minipage}{\textwidth}
\centering
\includegraphics[width=0.89\textwidth]{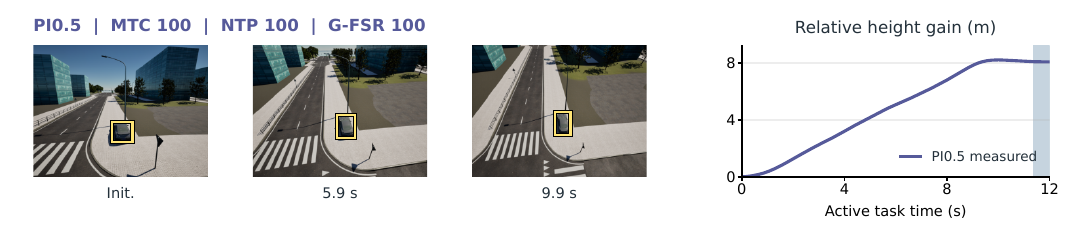}
\captionof{figure}{\textbf{E1 on CityRoads: Successful rise and hold.} PI0.5 elevates the camera by approximately $8$\,m relative to the stationary vehicle while keeping it cleanly in frame. Relative height remains stable during the final $0.6$\,s (shaded).}
\label{fig:focused_elevation}
\end{minipage}\par

\end{document}